%% file: main.tex
\documentclass[letterpaper]{article} 
\usepackage[draft]{aaai2026}  
\usepackage{times}  
\usepackage{helvet}  
\usepackage{courier}  
\usepackage[hyphens]{url}  
\usepackage{graphicx} 
\usepackage{natbib}  
\usepackage{caption} 
\usepackage{amsmath, amsfonts, amssymb}
\usepackage{mathtools}
\usepackage[noabbrev]{cleveref}
\usepackage{siunitx}
\usepackage{booktabs}
\usepackage{tikz}
\usetikzlibrary{shapes, positioning, calc, decorations,decorations.pathreplacing}
\usepackage{chemfig}
\usepackage{listings}
\usepackage{caption}
\usepackage{subcaption}
\usepackage{algorithm}
\usepackage{algpseudocode}

\usepackage{newfloat}
\usepackage{listings}
\DeclareCaptionStyle{ruled}{labelfont=normalfont,labelsep=colon,strut=off} 
\floatstyle{ruled}
\newfloat{listing}{tb}{lst}{}
\floatname{listing}{Listing}

\makeatletter
\AddToHook{cmd/appendix/before}{\def\cref@section@alias{appendix}}
\makeatother

\setchemfig{
    angle increment=30,
    atom sep=1.67em,
    double bond sep=0.67ex,
    bond style={line width=0.1em},
    cram width=0.8ex,
    cram dash width=0.1em,
    cram dash sep=0.2em,
    arrow style={line width=0.067em},
    arrow head=-{Triangle},
    arrow label sep=1ex,
    cycle radius coeff=0.75,
    chemfig style={line width=0.1em},
}
\setcharge{
    extra sep = 0.5ex,
    .radius = 0.05em,
    .style = {fill = fg},
}

\newcommand\mdoubleplus{\mathbin{+\mkern-5mu+}}

\DeclareMathOperator*{\argmin}{arg\,min}

\title{Diverse Mini-Batch Selection in Reinforcement Learning for Efficient Chemical Exploration in \textit{de novo} Drug Design}
\author{
    Hampus Gummesson Svensson\textsuperscript{\rm 1, \rm 2},
    Ola Engkvist\textsuperscript{\rm 1, \rm 2},
    Jon Paul Janet\textsuperscript{\rm 1},
    Christian Tyrchan\textsuperscript{\rm 3},
    Morteza Haghir Chehreghani\textsuperscript{\rm 2}
}
\affiliations{
    \textsuperscript{\rm 1}Molecular AI, Discovery Sciences, R\&D, AstraZeneca, Gothenburg, Sweden\\
    \textsuperscript{\rm 2}Department of Computer Science and Engineering, Chalmers University of Technology and University of Gothenburg, Gothenburg, Sweden\\
    \textsuperscript{\rm 3}Medicinal Chemistry, Research and Early Development, Respiratory and Immunology (R\&I), BioPharmaceuticals R\&D, AstraZeneca, Gothenburg, Sweden\\
    hamsven@chalmers.se
    
}

\begin{document}

\maketitle

\begin{abstract}
  In many real-world applications, evaluating the quality of instances is costly and time-consuming, e.g., human feedback and physics simulations, in contrast to proposing new instances. In particular, this is even more critical in reinforcement learning, since it relies on interactions with the environment (i.e., new instances) that must be evaluated to provide a reward signal for learning. At the same time, performing sufficient exploration is crucial in reinforcement learning to find high-rewarding solutions, meaning that the agent should observe and learn from a diverse set of experiences to find different solutions. Thus, we argue that learning from a diverse mini-batch of experiences can have a large impact on the exploration and help mitigate mode collapse.
  In this paper, we introduce mini-batch diversification for reinforcement learning and study this framework in the context of a real-world problem, namely, drug discovery. We extensively evaluate how our proposed framework can enhance the effectiveness of chemical exploration in \textit{de novo} drug design, where finding diverse and high-quality solutions is crucial. Our experiments demonstrate that our proposed diverse mini-batch selection framework can substantially enhance the diversity of solutions while maintaining high-quality solutions. In drug discovery, such an outcome can potentially lead to fulfilling unmet medical needs faster.
\end{abstract}

\section{Introduction}
In recent years, utilizing reinforcement learning (RL) for fine-tuning of pre-trained generative models has shown great success in various applications \cite{zhai2024fine,fan2023dpok}, including \textit{de novo} drug design \cite{olivecrona2017molecular,atance2022novo}. 
\textit{De novo} drug design is a computational problem that aims to identify novel molecular structures with specific properties without any starting template \cite{mouchlis2021advances}, where generative models have shown great success \cite{tong2021generative,pang2023deep}.
When fine-tuning a generative model, the goal is often to align the model's outputs with respect to human preferences or experiments. 
However, many practical applications require frequent assessment of data and experiences, e.g., via human expert evaluation, computer simulations, field testing, and laboratory experimentation. These assessment methods are often resource-intensive, demanding significant time and financial investment. In \textit{de novo} drug design, resource-intensive computational methods are used to assess the fit of molecules into the binding site of a target protein to predict the strength of each protein-ligand interaction \cite{paggi2024art}. Consequently, the volume of data that can undergo thorough evaluation is often constrained by budgetary limitations.

In this paper, we tackle this problem in reinforcement learning, where the training instances are provided solely from the agent's interaction with the environment. 
In particular, we study this problem in the context of \textit{de novo} drug design, where RL techniques are commonly used to fine-tune a pre-trained generative model to produce molecules with desired properties \cite{patronov2021has,pitt2025real}.
 In general, many successful RL algorithms, e.g., \cite{schulman2017proximal,mnih2016asynchronous}, run many copies of the environment in parallel to synchronously or asynchronously learn from numerous interactions. For synchronous on-policy algorithms, the experiences are accumulated to compute an average loss to update the agent's policy. This is also true for \textit{de novo} drug design \cite{olivecrona2017molecular}, where for each policy update, a batch of molecules is first generated in parallel. However, in many real-world applications, including \textit{de novo} drug design, it is impractical to assess all interactions with the environment, where each assessment of an interaction provides a reward signal for the agent. Instead, it is preferable to evaluate a smaller, representative set and learn from it. 

At the same time, to avoid mode collapse, exploration mechanisms play a vital role in agent performance, especially in tasks with delayed/sparse reward or for a reward landscape with a vast number of local optima to explore.
In \textit{de novo} drug design, a reward can only be obtained when the full molecular structure has been generated. Moreover, diversity among generated molecules is essential since a diverse molecular library increases the likelihood of identifying candidates with unique and favorable pharmacological profiles, thereby enhancing the overall efficiency and success rate of drug development pipelines. In drug design, the reward function is often complex and has many high-rewarding modes that should be found and subsequently exploited to obtain a diverse set of solutions. Thus, chemical exploration and diversification are of integral importance in drug design. In real-world deployment of this \textit{de novo} drug design, it is also often costly and time-consuming to evaluate an instance (i.e., a state-action episode) to obtain a reward. This creates a \emph{reward bottleneck} which limits the policy updates, leading to the need for efficient exploration.

One popular approach to enhance exploration in RL is the addition of an exploration bonus to the reward function, commonly denoted as \emph{intrinsic reward} \cite{burda2018exploration, Badia2020Never, seo2021state,NIPS2017_3a20f62a}. Another common approach is maximum entropy RL, where the agent tries to maximize both the reward and entropy simultaneously, i.e., succeeding at a task while still acting as randomly as possible \cite{o'donoghue2017combining,haarnoja2017reinforcement}. Our work provides a consistent perspective where, while improving exploration
by achieving diverse behaviors, it is important to make sure that the interactions with the environment are of high quality (i.e., receive high rewards). 
This becomes especially critical when the agent must account for safety considerations, exhibits sensitivity to noise, or operates in environments where numerous trajectories are infeasible. For example, in \textit{de novo} drug design, a molecular representation may not correspond to a chemically viable compound, and minor modifications can readily compromise its validity. In this work, we accomplish this by considering mini-batch diversification in reinforcement learning, where a large number of interactions (obtained from running copies of the environment in parallel) are summarized in a smaller, diverse set of interactions used for updating the policy. This provides an effective way to impose additional exploration in the learning process, while overcoming the reward bottleneck by learning from a smaller set.

In this paper, we argue that providing a diverse mini-batch of interactions makes the agent's exploration more effective and increases the diversity of the forthcoming interactions, especially in \textit{de novo} drug design. Thus, there are two key benefits for such mini-batch diversification: (1) computational aspects to address the reward bottleneck; (2) enhance exploration by diverse behaviors. Therefore, we introduce a framework for diverse mini-batch selection in reinforcement learning, which is illustrated in \cref{fig:process}.  
To the best of our knowledge, this is the first effort to study the effects of diverse mini-batch selection in reinforcement learning to overcome the reward bottleneck and promote exploration.
We study the use of determinantal point processes (DPP) \cite{kulesza2012determinantal}, the MaxMin algorithm \cite{ashton2002identification} and $k$-medoids clustering \cite{rdusseeun1987clustering} for this task. 
DPPs provide an effective framework to sample a diverse set based on specified similarity information, while the MaxMin algorithm and $k$-medoids clustering seek to choose a subset to maximize the coverage of a larger set. Previous work has proposed a mini-batch diversification scheme based on DPPs for stochastic gradient descent and shown its effectiveness \cite{zhang2017determinantal,huang2019efficient}, but such a scheme has not been applied to reinforcement learning. Also, previous work has used DPPs in diverse sampling for batch Bayesian Optimization \cite{nava2022diversified}. In this paper, we focus on mini-batch diversification for improving exploration and reducing reward computations (i.e., addressing the reward bottleneck) in reinforcement learning. In reinforcement learning, DPPs have previously been used for unsupervised option
discovery \cite{chen2023unified}, diverse recommendations for RL-based user preferences \cite{liu2021diversity}, and multi-agent RL \cite{sheikh2022dns,yang2020multi, osogami2019determinantal}. All of these are different from our setting and can not be applied to our setting. The MaxMin algorithm is a popular method used in drug discovery to pick a diverse set \cite{dreiman2021changing,tan2022drlinker}, but has not been investigated in combination with reinforcement learning. Furthermore, $k$-medoids clustering is a widely known clustering technique for finding a good partition in non-Euclidean data and has only been used for cluster-based RL \cite{grua2018exploring}, which is different from our setting.
To the best of our knowledge, our paper provides the first combinations of these methods with reinforcement learning to effectively fine-tune a generative model for \textit{de novo} drug design (or any other application). 

Thereby, the contribution of this paper is twofold: 
\begin{itemize}
    \item We propose a mini-batch diversification framework for RL to enhance exploration and, at the same time, to address the reward bottleneck issue.
    \item We extensively investigate the proposed framework on the \textit{de novo} drug design application, and demonstrate its effectiveness via extensive experiments. 
\end{itemize}
Due to the characteristics of the \textit{de novo} drug design problem, it is a suitable problem to employ diverse mini-batch selection and study its effectiveness.
We believe that this framework can also help to overcome the reward bottleneck and enhance exploration in other real-world applications of reinforcement learning, especially for fine-tuning a pre-trained generative model in other domains.  
Exploration is a key challenge in RL, and domain-specific information can easily be incorporated into the proposed framework.

\begin{figure}[t]
    \centering
    \resizebox{0.8\hsize}{!}{%
    \input{dpp_rl.tikz}
    }
    \caption{We propose a framework for diverse mini-batch selection in reinforcement learning. The RL agent generates a set of experiences in parallel, e.g., trajectories. A kernel measures the pairwise similarities between trajectories and is used to select a diverse set. The selected set is evaluated and, subsequently, is used to update the RL agent.}
    \label{fig:process}
\end{figure}

\section{Background}
\subsection{RL-based \textit{de novo} Drug Design}
The aim of \textit{de novo} drug design is to design novel drug molecules given a set of predefined constraints, but without any known initial structure \cite{mouchlis2021advances}. 
A popular approach for \textit{de novo} drug design is to use chemical language models to generate string-based representations of molecules \cite{arus2019exploring,segler2018generating}. To steer the chemical language model to promising areas of the chemical space, reinforcement learning can be leveraged \cite{olivecrona2017molecular}. This paper focuses on promoting diversity in RL-based fine-tuning of a chemical language model via mini-batch diversification.
An action $a$ in this RL problem corresponds to adding one token to the string representation of the molecule, where $\mathcal{A}$ is the set of possible tokens that can be added, including a start token $a^{\text{start}}$ and a stop token $a^{\text{stop}}$. The reward function assesses the quality of the molecule represented by the string, and the molecule can only obtain a reward when the full string representation has been generated, i.e., a stop token has been added. This \textit{de novo} drug design problem can be modeled as a Markov decision process (MDP), e.g., see \cite{gummesson2024utilizing} for more details.

One popular string-based representation of chemical entities is Simplified Molecular Input Line Entry System \cite{weininger1988smiles}, abbreviated SMILES. Evaluations by both \citet{gao2022sample} and \citet{thomas2022re} have concluded good performance of the SMILES-based REINVENT \cite{segler2018generating, olivecrona2017molecular,blaschke2020reinvent,loeffler2024reinvent} compared to both other RL-based and non-RL-based approaches for \textit{de novo} drug design. REINVENT consists of a long short-term memory (LSTM) network \cite{hochreiter1997long} using SMILES to represent molecules as text strings. REINVENT utilizes an on-policy RL algorithm to perform online optimization of the policy $\pi_\theta$ to generate higher-rewarding molecules. Previous work has shown that minimizing its loss function is equivalent to maximizing the expected return, as for policy gradient algorithms \cite{guo2024augmented}. Our work builds upon the success of REINVENT and focuses on improving its chemical exploration and avoiding mode collapse.

\subsection{Diversity in \textit{de novo} drug design}
The drug-like chemical space is estimated to consist of $10^{33}$ synthesisable molecules \cite{polishchuk2013estimation}. To explore this space and improve the diversity of the generated molecules, several studies aim to improve the chemical exploration carried out by the RL agent. Without the use of any exploration technique, the policy easily collapses to generating only a few modes of the reward function, which leads to low diversity. To improve the diversity in RL-based \textit{de novo} drug design, \citet{blaschke2020memory} therefore introduces a count-based method that reduces the reward for similar molecules based on their structure. The work of \citet{park2024molairmolecularreinforcementlearning} and \citet{wang2024exselfrl} employs memory and learning-based intrinsic motivation to improve the reward of the generated molecules. Moreover, previous work shows that incorporating both structure- and learning-based information into the reward function can improve the overall diversity of \textit{de novo} drug design \cite{svensson2024diversityawarereinforcementlearningnovo}. Our work takes on a fundamentally alternative perspective to enhance diversity. Rather than just encouraging diverse and explorative behavior via the reward signal, our work studies the effect of maximizing the diversity of the molecules that we evaluate and learn from. 

 To measure the diversity among a given set of molecules, several existing metrics have been proposed. \citet{hu2024hamiltonian} divides these metrics into two main categories: reference-based and distance-based. A reference-based metric compares a molecular set with a reference set to find the intersection. Distance-based metrics use pairwise distances among the molecular set to determine the diversity. 
In this work, both metrics are applied. As the representative reference-based metric, the number of molecular scaffolds, also known as Bemis-Murcko scaffolds \cite{bemis1996properties}, is used. As a distance-based metric, we utilize the number of \emph{diverse actives}\footnote{Diverse actives is termed diverse hits in previous work by \citet{renz2024diverse}.} metric by \citet{renz2024diverse}, which is based on \#Circles metric proposed by \citet{xie2023much}. Following the definition by \citet{renz2024diverse} but using the terminology of predicted active molecules (rather than hit molecules), the number of diverse actives for distance threshold $D$ is defined by
\begin{equation}
\label{eq:diverse_active}
\resizebox{0.89\hsize}{!}{%
    $\mu\left(\mathcal{H}; D\right) = \max_{\mathcal{C}\in 2^\mathcal{H}} |\mathcal{C}|\text{ s.t. } \forall x \neq y \in \mathcal{C}: d(x,y) \geq D,$ %
    }
\end{equation}
where $\mathcal{H}$ is a set of predicted active molecules, $2^\mathcal{H}$ is the power set, $d(x,y)$ is the distance between molecules $x$ and $y$. As suggested by \cite{renz2024diverse}, we use the MaxMin algorithm \cite{ashton2002identification} implemented in RDKit \cite{landrum2006rdkit} to find an approximate maximal value of the cardinality of $C$.

\section{Diverse Mini-Batch Selection For RL}
\begin{algorithm}
\caption{Diverse Mini-Batch Selection}\label{alg:framework} 
\begin{algorithmic}[1] 
\State \textbf{input:} $G,B,k, \theta_0, T, p_0$
\State $\mathcal{M} \gets \emptyset$ 
\State $\theta \gets \theta_0$ \Comment{Initial policy parameters}
 \For{$g=1,\dots,G$}
    \For{$b=1,\dots,B$} \Comment{Generate in parallel}
    \State $s_0 \sim p_0(\cdot)$ \Comment{Sample first state}
    \For{$t=0,1,\dots,T-1$}
        \State $a_t \sim \pi_\theta(s_t)$
        \State Observe next state $s_{t+1} \sim P(\cdot|s_t, a_t)$
    \EndFor
    \State $\tau_{b} := s_0,a_0,\dots,a_{T-1},s_T$ 
    \Comment{Trajectory}
    \EndFor
    \State $\mathcal{B} \gets \{\tau_1,\dots,\tau_B\}$

    \State Compute kernel matrix $L$ over $\mathcal{B}$
    \State Select $k$ representative trajectories from $\mathcal{B}$ 
    \State $\forall \tau \in Y,$ observe return $r(\tau)$  \Comment{Evaluation}
    \State $\mathcal{M} \gets \mathcal{M} \cup \left(\cup_{\tau\in Y}\{\tau,r(\tau)\}\right)$ 
    \State Update $\theta$ using RL algorithm
\EndFor
 \State \textbf{output:} $\theta, \mathcal{M}$
\end{algorithmic} 
\end{algorithm}
We propose a framework to enhance exploration in reinforcement learning while reducing the number of interactions evaluated. We seek to generate more diverse solutions through reinforcement learning-based fine-tuning of a pre-trained generative model. In this paper, we focus on fine-tuning a chemical language model. We assume delayed rewards and that acquiring a sequence of states and actions is inexpensive compared to the evaluation, which is often true for real-world problems such as \textit{de novo} drug design. Given a large set of interactions, we seek to select a smaller, representative set to use for updating the parameters of our policy. 
We hypothesize that this affects the agent's exploration of the solution space, which is of vast importance in RL-based \textit{de novo} drug design, while overcoming the reward bottleneck by considering a fixed budget of evaluations. 
The intuition is that learning from diverse experiences helps the agent to explore more effectively. 

Therefore, we suggest enforcing diversity among the selected interactions to improve the efficiency of the exploration. For this purpose, we propose a diverse mini-batch selection framework for reinforcement learning, which is illustrated in \cref{alg:framework}. Here, we focus on trajectories of actions, i.e., an interaction corresponds to a trajectory, which we use as a more general notion of an episode. However, the framework can easily be extended beyond trajectories/episodes. In the \textit{de novo} drug design problem that we consider, an episode corresponds to a fully generated molecule, since each action in the episode corresponds to adding a character in the SMILES representation. Also, we consider policy-based RL, where we directly learn a policy $\pi_\theta$ with policy parameters $\theta$, but the suggested framework can also be applied to value-based RL algorithms, e.g., by diverse mini-batch selection from the replay buffer.

Over $G$ training/generative steps, using the agent's current policy $\pi_\theta$, a batch $\mathcal{B}$ of $B$ trajectories is sampled in parallel over copies of the same environment. Each trajectory has a maximum horizon of $T$ steps, where the true length of each trajectory can depend on some stopping criteria or when the terminal state is reached. If $B$ is chosen such that $B \gg k$ and the agent's policy is stochastic, this set will contain primarily unique items. We let the RL agent in each copy of the environment focus on maximizing the expected return of each trajectory with respect to the reward function, i.e., maximizing the return generated by the agent's policy 
\begin{equation}
   \mathbb{E}_{\tau\sim \pi_\theta} [R(\tau)],
\end{equation}
where $\tau$ is a state-action trajectory $S_0,A_0,S_1\dots,S_{T-1},A_{T-1},S_T$, $R(\tau)$ is the return of following $\tau$ and $\mathbb{E}_{\pi_\theta}[\cdot]$ denotes the expected value of a random variable given that the agent follows policy $\pi_\theta$. This generates $\mathcal{B}$ under the belief that the agent tries to maximize each return, without explicitly considering the diversity among individual trajectories. This can be particularly important when the agent has to consider safety concerns, is sensitive to noise, or when many trajectories are not viable. For instance, in \textit{de novo} drug design, a SMILES string is not necessarily chemically feasible, and small changes can easily break its validity. Therefore, it is important that the agent primarily focuses on generating chemically valid SMILES strings of high quality. Moreover, the proposed method can be combined with other exploration techniques, e.g., intrinsic motivation \cite{burda2018exploration,NIPS2017_3a20f62a}, to provide additional domain-specific exploration. Given a large batch of trajectories $\mathcal{B}$, to stay within the given budget of evaluations (per generative step), the next step is to obtain a smaller, diverse mini-batch $Y$ that summarizes $\mathcal{B}$. We study the use of determinantal point processes (DPPs) \cite{kulesza2012determinantal}, the MaxMin algorithm \cite{ashton2002identification} and $k$-medoids clustering \cite{rdusseeun1987clustering} for this task. After a set $Y$ of $k$ trajectories has been obtained, each trajectory in $Y$ is evaluated to obtain the corresponding returns and/or state-action rewards. Using the returns and rewards, the policy parameters are updated by employing an arbitrary RL algorithm. The discussed framework is agnostic to the RL algorithm used to update the policy parameters, and yields both the policy parameters $\theta$ and a diverse set of trajectory-return pairs $\{\tau, R(\tau)\}$.

\subsection{Determinantal Point Processes (DPPs)}
We propose and study the use of determinantal point processes \cite{kulesza2012determinantal} to sample a diverse mini-batch for RL updates. DPPs provide an effective framework to sample a diverse set based on specified similarity information. To the best of our knowledge, our work is a novel combination of DPP and reinforcement learning to effectively fine-tune a generative model.

A point process $\mathcal{P}$ is a probability measure over finite subsets of a set $\mathcal{B}$. We consider the discrete case of $\mathcal{B} = \{1,2,\dots, B\}$, where $B$ is the number of unique trajectories. In this case, a point process is a probability measure on the power set $2^{\mathcal{B}}$, i.e., the set of all subsets of $\mathcal{B}$. Determinantal point processes (DPPs) are a family of point processes characterized by the \emph{repulsion} of items such that similar items are less likely to co-occur in the same sample. Given a kernel, providing a similarity measure between pairs of items, DPP places a high probability on subsets that are diverse with respect to the kernel. We consider a class of DPPs named L-ensembles \cite{borodin2005eynard}, which is defined via a real, symmetric matrix $L$ over the entire (finite) domain of $\mathcal{B}$. This matrix is often denoted as the \emph{kernel matrix}. The probability of subset $Y \subseteq \mathcal{B}$ is given by
\begin{equation}
    \mathcal{P}_L (Y) \propto \det (L_Y),
\end{equation}
where $L_Y = [L_{ij}]_{i,j \in Y}$ denotes the restriction of $L$ to the entries indexed by items of $Y$. Thus, the probability of sampling the set $Y \subseteq \mathcal{B}$ is proportional to the determinant of $L_Y$ restricted to $Y$. The normalization constant is available in closed form since $\sum_{Y\subseteq \mathcal{B}} \det (L_Y) = \det (L+ I)$, where $I$ is the $N\times N$ identity matrix. 

Given the larger set $\mathcal{B}$, we want the smaller set $Y$ to contain a pre-defined number of items from $\mathcal{B}$. Thus, we are interested in sampling a subset $Y$ with a fixed cardinality $|Y| = k$ to sample a mini-batch with a fixed size. $k$-DPPs \cite{kulesza2011k} concern DPPs conditioned on the cardinality of the random subset. Formally, the probability of a $k$-DPP to sample a subset $Y \subseteq \mathcal{B}$ is given by 
\begin{equation}
    \mathcal{P}_L^k (Y) = \frac{\det (L_Y)}{\sum_{Y'\subseteq \mathcal{B}:|Y'|=k}\det (L_{Y'})},
\end{equation}
where $|Y| = k$. The $k$-DPP's inherent ability to promote diversity makes it an excellent choice for selecting diverse and representative mini-batches in reinforcement learning. In this way, $k$-DPP provides a smaller and diverse set of items from a larger set of items. 

How the $k$-DPP will summarize the larger set is determined by the kernel matrix $L$. Constructing the kernel matrix entails using domain knowledge, but other information can also be used. 
Let $q_i \in \mathbb{R}^+$ be a quality term and $\phi_i \in \mathbb{R}^D$, $||\phi_i|| = 1$, a vector of normalized diversity features of the $i$-th item in $\mathcal{B}$, e.g., the $i$-th generated SMILES string. Following the work of \citet{kulesza2012determinantal}, the entries of the kernel matrix can then be expressed 
\begin{equation}
    L_{ij} = q_i \phi_i^T\phi_j q_j,
\end{equation}
where $q_i$ is a quality term measuring the intrinsic ``goodness'' of the $i$-th item, and $\phi_i^T \phi_j \in [-1,1]$ is a signed measure of similarity between $i$-th and $j$-th item.
Therefore, utilizing $k$-DPPs allows for a flexible sampling procedure that behaves differently depending on the information incorporated in the kernel matrix $L$. It does not directly optimize the determinant of $L$, but instead includes randomness to encourage additional exploration. In the \textit{de novo} drug design problem studied in this paper, we only consider the similarity between items and do not explore the effects of quality terms. The reason for this is that we focus on pure diversification, and we assume that the items generated by the policy have similar quality. Our preliminary studies on \textit{de novo} drug design did not find any performance gain in incorporating a quality term provided by an oracle. However, we believe that it can be beneficial to include a quality term, but different terms need to be investigated to find a suitable one.

\subsection{Maximum Coverage}
As an alternative to selecting a representative set by sampling via $k$-DPPs, we also study mini-batch diversification by maximizing the coverage of the larger set for a fixed cardinality. While $k$-DPPs provide a sampling procedure to summarize a larger set given a kernel matrix, maximum coverage aims to directly cover as large a part of the space as possible. In this way, we seek to pick the most diverse items subject to the cardinality constraint of $k$. For a given set $\mathcal{B}$ of $B$ candidate items, let $f(Y)$ be a function that measures the ``coverage'' of any given set $Y$ of items. The goal is to choose a set $Y$ of $k$ items such that $f(Y)$ is maximized. Here we consider a fixed size of $k$, but a possible extension could be to choose the smallest set $Y$ such that a sufficient coverage of $\mathcal{B}$ is obtained. Formally, we define this problem by 
\begin{equation}
    \max_{Y\in [\mathcal{B}]^k} f(Y),
\end{equation}
where $[\mathcal{B}]^k \triangleq \{X \in 2^{\mathcal{B}} : |X| = k\}$ is the set of all subsets with cardinality $k$. In this work, we consider coverage functions $f(Y)$ based on dissimilarities between trajectories. 

We investigate two algorithms to find an approximate maximum coverage of the large set: (1) the MaxMin algorithm \cite{ashton2002identification}, implemented by RDKit \cite{landrum2006rdkit}; (2) $k$-medoids clustering \cite{rdusseeun1987clustering}, using the FasterPAM algorithm \cite{schubert2019faster,schubert2021fast} implemented by \citet{schubert2022fast}.
The MaxMin algorithm first picks a starting item, creating a picked set. Then the algorithm iteratively, from the items in the candidate pool, finds the item that has the maximum dissimilarity to molecules in the picked set and adds this item to the picked set. The MaxMin algorithm is widely used in drug discovery to pick a diverse set \cite{dreiman2021changing,tan2022drlinker}. 

$k$-medoids clustering \cite{rdusseeun1987clustering} is a popular technique to cluster non-Euclidean data using arbitrary dissimilarities or input domains. The k-medoids problem aims to split $B$ items into $k$ ($\leq B$) clusters, where the number of clusters is assumed to be specified beforehand. The medoid of a cluster is defined as the item in the cluster with the minimum average of dissimilarity to all the other items in the cluster, i.e., the item that is most centrally located within the cluster. Unlike several other clustering algorithms, e.g., $k$-means \cite{arthur2007k}, the medoid is an actual item in the cluster. Thus, the objective is to find medoids $m_1,\dots,m_k$ that minimizes
\begin{equation}
    \argmin_{\{m_1,\dots,m_k\} \subset Y} \sum_{i=1}^k \sum_{x_c\in C_i} d(x_c,m_i), 
\end{equation}
where $C_i$ is the cluster of medoid $m_i$ and $d$ is an arbitrary dissimilarity function. While the MaxMin algorithm sequentially adds items to the picked set in a greedy manner, $k$-medoids simultaneously seeks to optimize all medoids to find the best picks. Finding the global optimum of the $k$-medoid problem is NP-hard \cite{kariv1979algorithmic}. Instead, the Partitioning Around Medoids (PAM) algorithm \cite{rdusseeun1987clustering}, which is the standard algorithm for $k$-medoids clustering, improves the clustering towards a local optimum. In this paper, we use the FasterPAM algorithm \cite{schubert2019faster,schubert2021fast}, which achieves a speedup in runtime compared to the original PAM algorithm, to select $k$ items (given by the medoids found by the algorithm).

\section{Experimental Evaluation}
We extensively evaluate our framework on \textit{de novo} drug design. We run experiments on three reward functions based on well-established molecule binary bioactivity label optimization tasks: the Dopamine Receptor D2 (DRD2), c-Jun N-terminal Kinases-3 (JNK3), and Glycogen Synthase Kinase 3 Beta (GSK3$\beta$) predictive activity models \cite{olivecrona2017molecular,li2018multi} provided by Therapeutics Data Commons \cite{Velez-Arce2024tdc}. The final (extrinsic) reward also includes the quantitative estimation of drug-likeness (QED) \cite{bickerton2012quantifying}, molecular weight, number of hydrogen bond donors, and structural constraints. For full specifications on the reward functions, we refer to \cref{app:experimental_details}.

To update the policy and generate SMILES, we use the REINVENT framework \cite{loeffler2024reinvent} with its pre-trained policy on the ChEMBL database \cite{gaulton2017chembl} to generate drug-like bioactive molecules. Previous benchmarks on \textit{de novo} drug design have, for this framework, concluded among the best performances \cite{gao2022sample,thomas2022re}, while it is also used in real-world applications \cite{pitt2025real}. The action space $\mathcal{A}$ consists of 34 tokens, including start and stop tokens. We evaluate the diversity of the generated set $\mathcal{M}$ by the number of molecular scaffolds and the number of diverse actives (see \cref{eq:diverse_active}), where the diverse actives are computed for every 250th generative step. For the diverse actives, we use Tanimoto dissimilarity to measure the distance between 2048-bit Morgan fingerprints (with radius 2 and computed by RDKit \cite{landrum2006rdkit}) and the distance threshold $D=0.7$ proposed by \cite{renz2024diverse}. When computing the diversity in terms of both scaffolds and diverse actives, we only regard active molecules, defined as molecules with both QED and predicted activity larger than $0.5$.

We compare the use of mini-batch diversification in combination with different techniques to modify the original reward for \textit{de novo} drug design: (1) no modification of the reward, i.e., the agent observes the original (extrinsic) reward; (2) using the popular identical molecular scaffold (IMS) penalty \cite{blaschke2020memory}, which sets the reward to $0$ when $M$ molecules with the same molecular scaffold have been generated; (3) using the TanhRND technique \cite{svensson2024diversityawarereinforcementlearningnovo}, which shows promising empirical results in terms of diversity. No modification of the reward is included as a baseline to investigate if mini-batch diversification can act as an alternative approach to avoid mode collapse by modifying the original reward. We hereafter denote the original reward without any modification as the \emph{extrinsic reward}. For mini-batch diversification with a mini-batch of $k=64$ SMILES, we first generate $B = 640$ SMILES via multinomial sampling and then use $k$-DPP to select a diverse mini-batch. Without mini-batch diversification, we directly generate $k=64$ SMILES via multinomial sampling, which is the standard procedure of the REINVENT framework. We denote these approaches without mini-batch diversification as \emph{diversification-free}.

\subsection{Construction of Kernel Matrix}
All of the investigated methods for mini-batch diversification (i.e., DPP, the MaxMin algorithm and $k$-medoids clustering) rely on a kernel matrix $L$ to encode the similarity between different molecules. We construct this kernel matrix based on two other kernel matrices $L_T$ and $L_D$, which we denote as ``base'' kernel matrices. The first base kernel matrix $L_T$ is constructed by the Tanimoto similarity between the corresponding 2048-bit Morgan fingerprints (with radius 2 using RDKit \cite{landrum2006rdkit}) of the generated SMILES. To incorporate more scaffold-based information, we construct the base kernel matrix $L_D$ by computing the Dice coefficients \cite{dice1945measures,sorensen1948method} between the scaffolds' atom pair fingerprints \cite{carhart1985atom}. Given these base kernels, we aggregate these base kernel matrices to define the kernel matrix $L$, which is used for selecting $k$ molecules, by $L = L_T + L_D$.
In \cref{app:kernel_matrix_dpp,app:kernel_matrix_max_cov}, we provide a study on different combinations of the base matrices to define $L$ and argue that the kernel matrix defined here provides the best balance between the different diversity metrics.

\subsection{Effects on Quality of Diverse Mini-Batch Selection}
\begin{figure}[ht]
     \centering
     \begin{subfigure}[b]{0.47\textwidth}
         \centering
         \includegraphics[width=\textwidth]{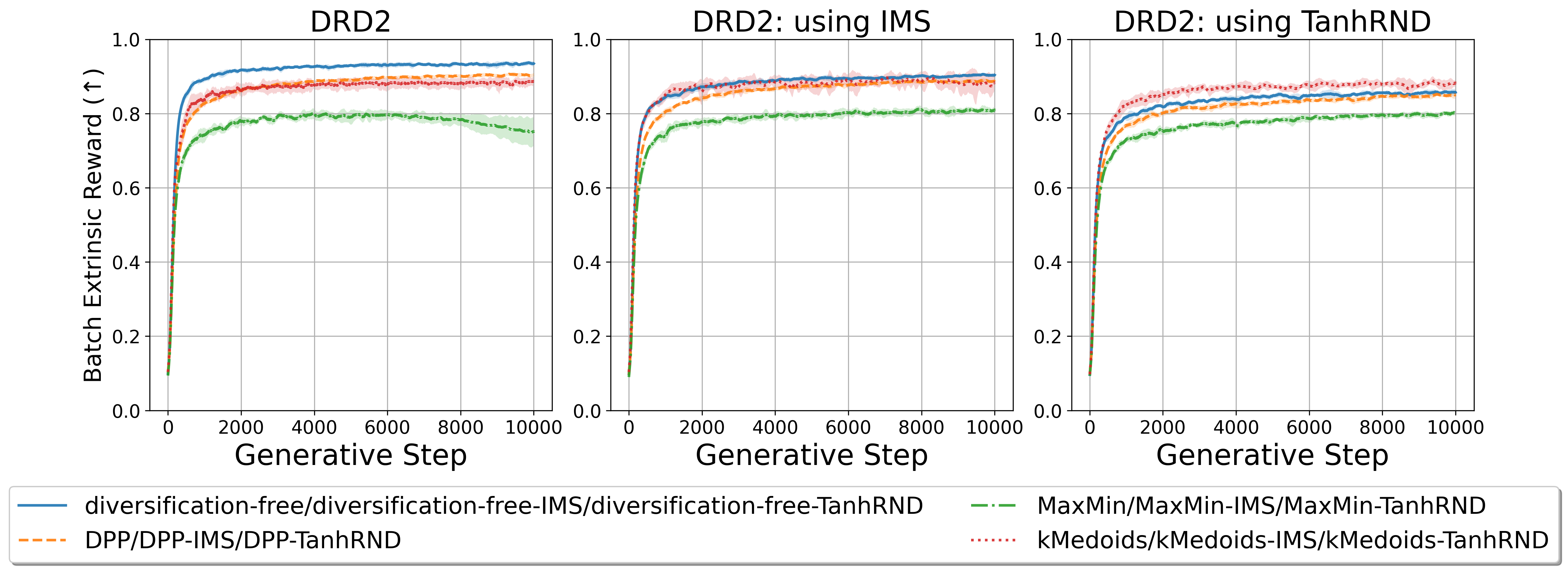}
         \caption{DRD2}
         \label{fig:drd2_reward}
     \end{subfigure}
     \hfill
     \begin{subfigure}[b]{0.47\textwidth}
         \centering
         \includegraphics[width=\textwidth]{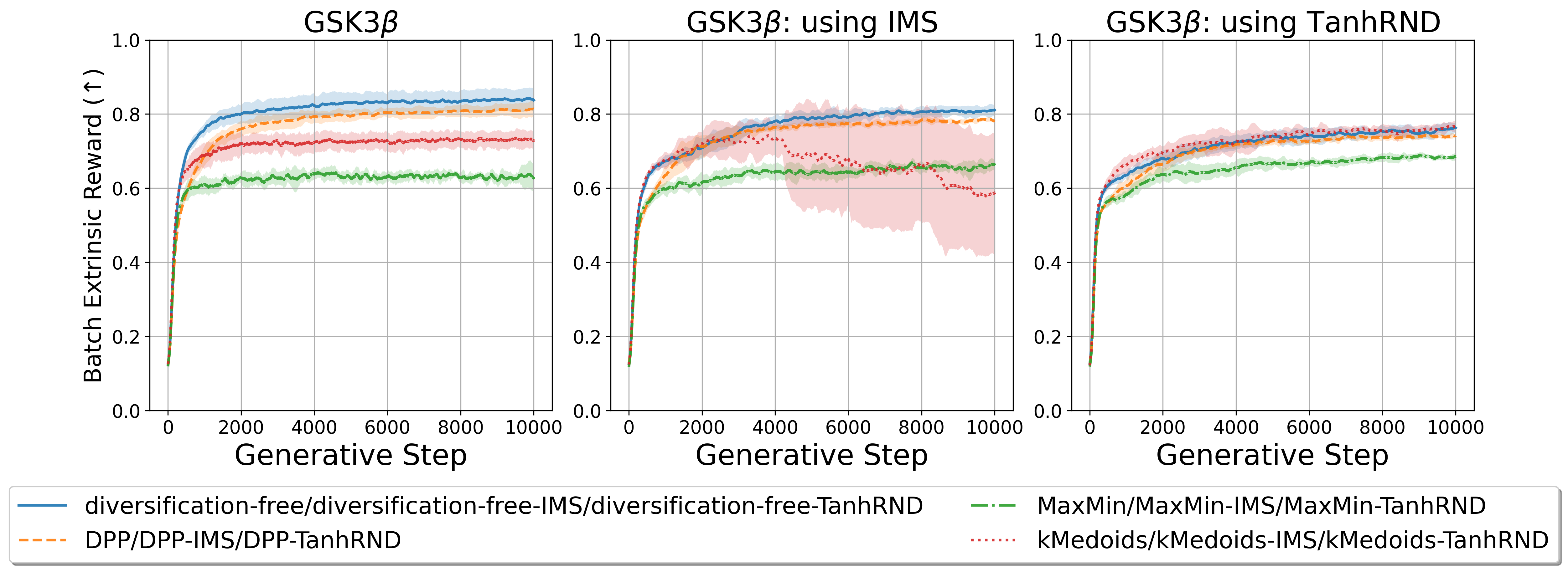}
         \caption{GSK3$\beta$}
         \label{fig:gsk3b_reward}
     \end{subfigure}
     \hfill
     \begin{subfigure}[b]{0.47\textwidth}
         \centering
         \includegraphics[width=\textwidth]{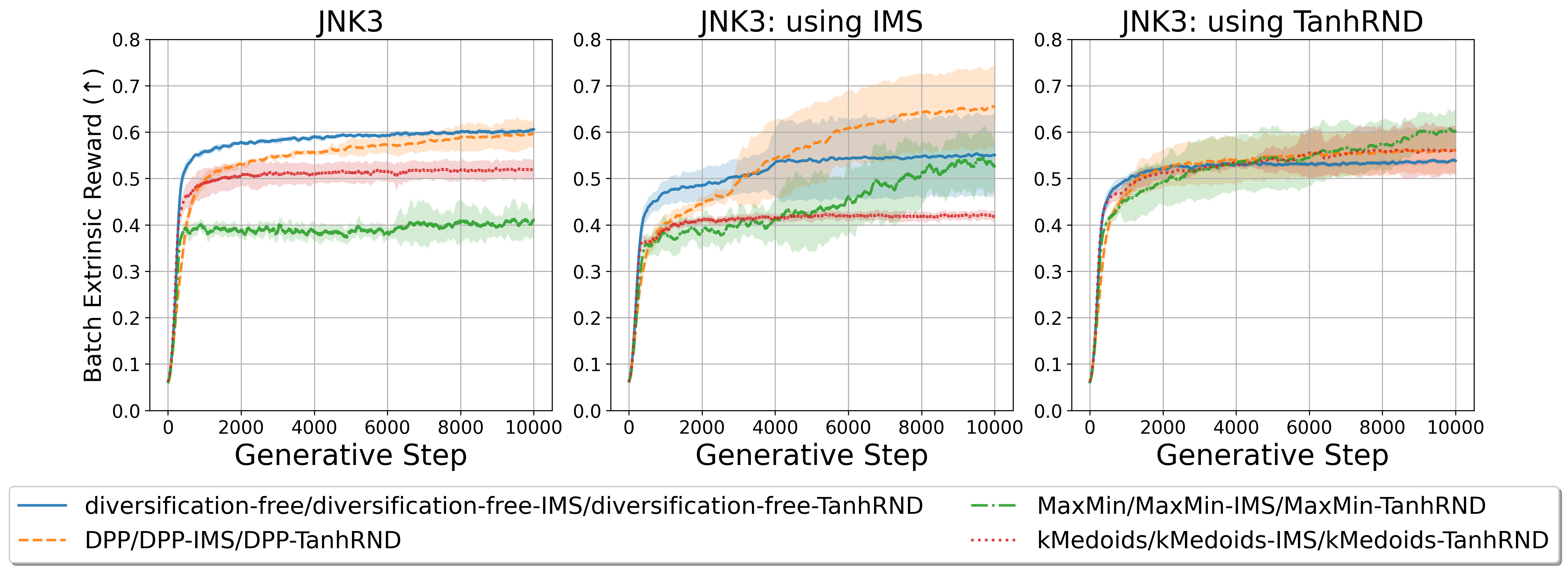}
         \caption{JNK3}
         \label{fig:jnk3_reward}
     \end{subfigure}
        \caption{Average extrinsic rewards per generative step across the mini-batch of SMILES evaluated on the DRD2-, GSK3$\beta$-, or JNK3-based reward functions. For clarity of presentation, we display the moving averages with a window size of 101. The average across 10 independent runs per generative step is plotted over \num{10000} generative steps, where the shaded area shows standard deviations among the independent runs. }
        \label{fig:reward}
\end{figure}
We first assess the quality (i.e., the reward) of the generated molecules to evaluate if our proposed framework can maintain high quality while enhancing the diversity.
Therefore, we study the extrinsic reward of each configuration. The extrinsic reward is the original reward provided for each molecule that we want to maximize, but not the reward observed by the agent when using IMS or TanhRND.

\Cref{fig:reward} displays the average extrinsic rewards for each mini-batch $Y$ of SMILES evaluated in each generative step. The average across 10 independent runs per generative step is plotted over \num{10000} generative steps, where the shaded area shows the corresponding standard deviation across the independent runs. For clarity of presentation, we show the moving averages with a window size of 101 (i.e., the current step and upto 50 steps on each side).
Each plot of \cref{fig:drd2_reward,fig:gsk3b_reward,fig:jnk3_reward} compares the use of diverse mini-batch selection using $k$-DPP, the MaxMin algorithm and $k$-medoids clustering in combination with different techniques of modifying the extrinsic (original) reward for \textit{de novo} drug design. The left plots compare the extrinsic rewards for both with and without mini-batch diversification when the extrinsic reward is not modified. The middle plots compare the extrinsic rewards when using the identical molecular scaffold (IMS) penalty proposed by \cite{blaschke2020memory} and the right plots display the comparisons when utilizing the TanhRND technique \cite{svensson2024diversityawarereinforcementlearningnovo}.

For the DPP and diversification-free methods on the DRD2- and GSK3$\beta$-based reward functions (see \cref{fig:drd2_reward,fig:gsk3b_reward}), we observe similar trends in terms of extrinsic reward, especially when using IMS or TanhRND. Moreover, on the DRD2 reward, these experiments achieve a reward of 0.8 or higher, while rewards close to 0.8 are achieved on the GSK3$\beta$ function. The diversification-free experiments converge faster, but the DPP experiments often converge to a similar average reward. Faster convergence tends to indicate that less exploration is performed, which is demonstrated in \cref{fig:circles,fig:scaffolds} below in terms of diversity of the generated molecules. $k$-medoids shows similar results on DRD2, but achieves more unstable and lower quality on GSK3$\beta$. For the MaxMin experiments on the DRD2 and GSK3$\beta$ problems, we observe that extrinsic rewards are lower than for both the DPP and diversification-free experiments. This is possibly because more exploration is enforced, due to a more diverse mini-batch, at the cost of less exploitation. For the experiments on the JNK3-based reward function (see \cref{fig:jnk3_reward}), we observe similar trends as for DRD2 and GSK3$\beta$ when not modifying the extrinsic reward (see left plot in \cref{fig:jnk3_reward}). On the other hand, when using the IMS or TanhRND technique to modify the extrinsic reward, all methods display similar extrinsic reward, but different convergence rates. Only $k$-medoids utilizing IMS performs differently, displaying an early convergence to a reward of around 0.4, which is lower than the other methods. This is likely due to insufficient exploration induced by this configuration. In general, the extrinsic rewards are significantly lower on JNK3, indicating that the JNK3-based reward function is more challenging to optimize. 
One possible explanation is that there are fewer active molecules for JNK3 in the ChEMBL database. When we evaluate molecules from ChEMBL on the DRD2, GSK3$\beta$, and JNK3 oracles, we observe that 2.4\%, 1.8\%, and 0.3\% of the molecules, respectively, have an oracle score above 0.5 (we refer to \cref{app:analysis_activity_models} for more details). Thus, there are fewer good solutions for JNK3. Since we use a model pre-trained on ChEMBL data, which limits the generation to molecules similar to those found in this data, the initial model is less likely to find sufficient solutions for JNK3.

\subsection{Diverse Mini-Batch Selection Enhances Distance-Based Diversity}
\begin{figure}[t]
     \centering
     \begin{subfigure}[b]{0.47\textwidth}
         \centering
         \includegraphics[width=\textwidth]{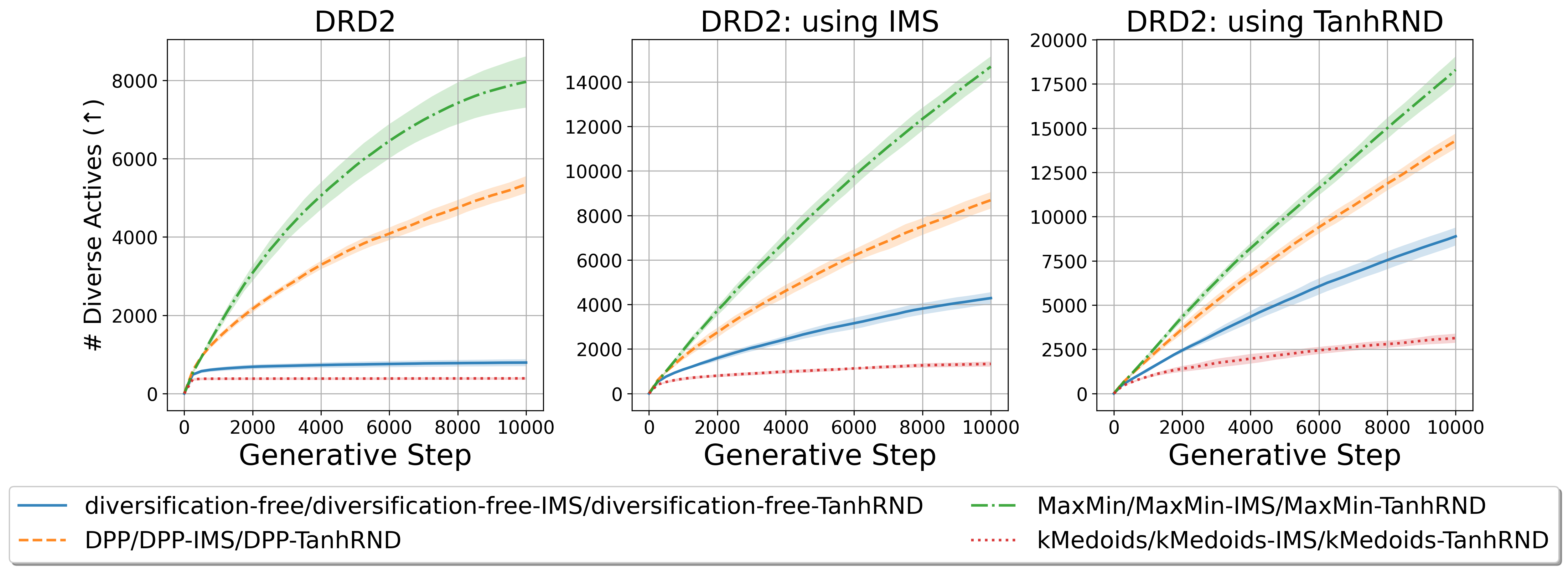}
         \caption{DRD2}
         \label{fig:drd2_circles}
     \end{subfigure}
     \hfill
     \begin{subfigure}[b]{0.47\textwidth}
         \centering
         \includegraphics[width=\textwidth]{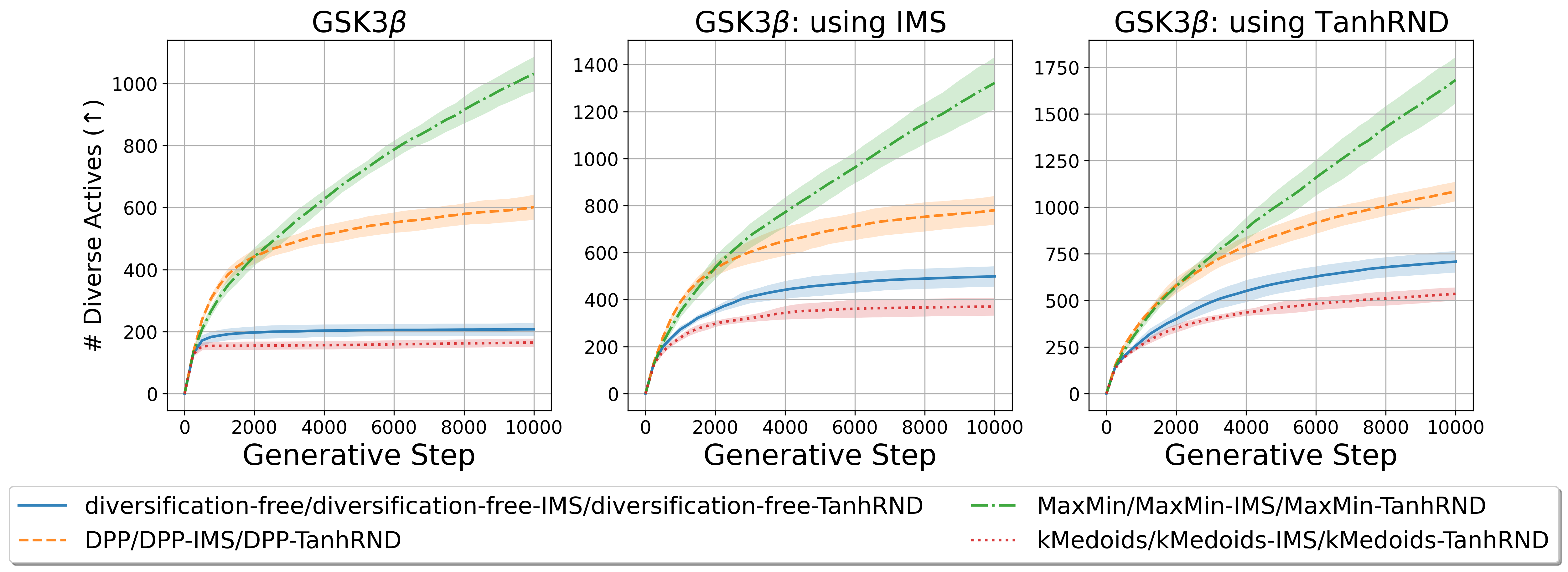}
         \caption{GSK3$\beta$}
         \label{fig:gsk3b_circles}
     \end{subfigure}
     \hfill
     \begin{subfigure}[b]{0.47\textwidth}
         \centering
         \includegraphics[width=\textwidth]{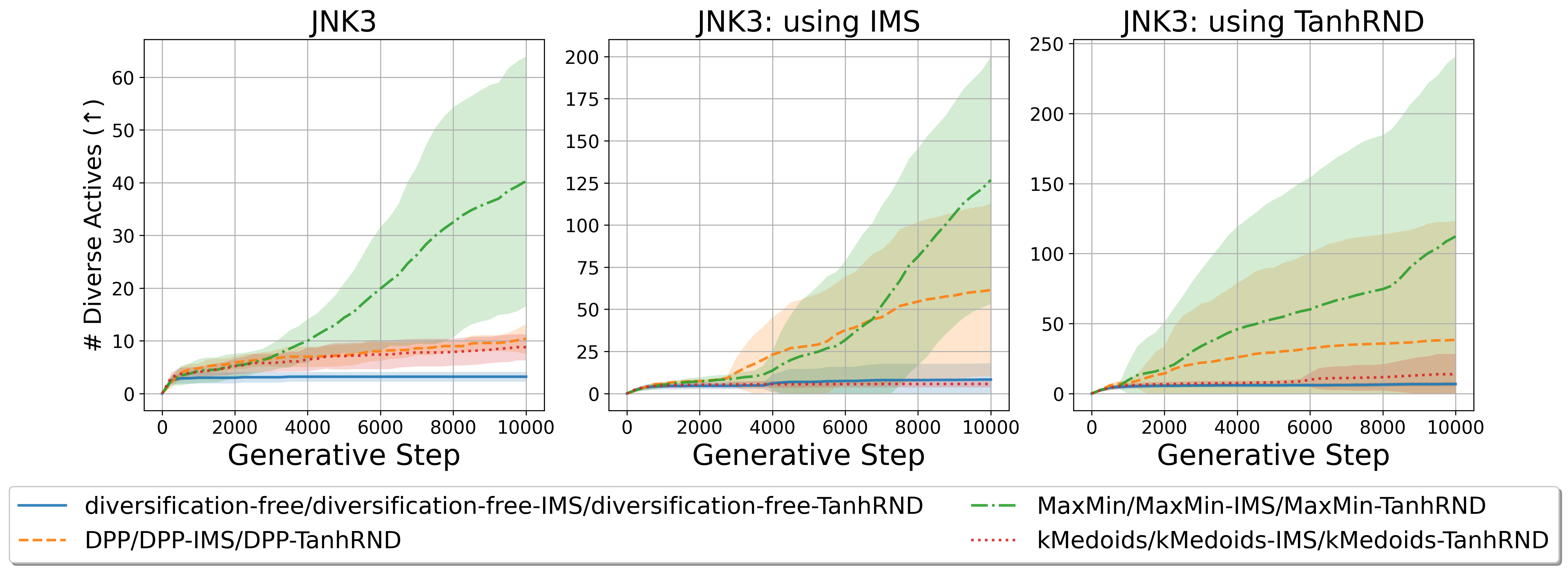}
         \caption{JNK3}
         \label{fig:jnk3_circles}
     \end{subfigure}
        \caption{Total number of diverse activities after $g$ generative steps evaluated on reward functions based on the DRD2, GSK3$\beta$, or JNK3 predictive model. The total number of diverse actives is plotted for every 250th generative step. The average across 10 independent runs per generative step is plotted over \num{10000} generative steps, where the shaded area shows standard deviations among the independent runs. } 
        \label{fig:circles}
\end{figure}
To evaluate the distance-based diversity among the generated molecules, we calculate the number of diverse actives. \Cref{fig:circles} shows the total number of diverse actives for every 250th generative step in the \textit{de novo} drug design task for the DRD2-, GSK3$\beta$- and JNK3-based reward functions. The lines and shaded area display the mean and standard deviation, respectively, across 10 independent reruns for each configuration. Each plot of \cref{fig:drd2_circles,fig:gsk3b_circles,fig:jnk3_circles} compares the use of DPP in combination with different techniques of modifying the extrinsic reward for \textit{de novo} drug design. 

\subsubsection{Dopamine Receptor D2 (DRD2)}
\Cref{fig:drd2_circles} displays the cumulative number of diverse actives per generative step on the DRD2-based reward function. We observe that utilizing mini-batch diversification significantly improves the total number of diverse actives found over 10000 generative steps compared to the diversification-free experiments (blue lines). We observe a significant gain after just a few hundred generative steps. In particular, MaxMin consistently yields the best results in terms of diverse actives, compared to the 

Interestingly, when not using IMS or TanhRND to modify the extrinsic reward (see left plot in \cref{fig:drd2_circles}), DPP and MaxMin display a considerable increase in distance-based diversity after a few hundred generative steps compared to the diversification-free method, where diversity quickly stagnates. Without mini-batch diversification (and any extrinsic reward modification), it is expected that the diversity should stagnate since it has previously been observed that the agent can easily get stuck in a local optimum and will then generate similar molecules \cite{blaschke2020memory}. Using mini-batch diversification via DPP or MaxMin overcomes this issue even without modifying the extrinsic reward, which is the standard method for tackling this issue. In addition, we observe that mini-batch diversification in combination with a modification of the extrinsic reward (see the middle and right plot in \cref{fig:drd2_circles}) yields the largest number of diverse actives, especially when utilizing TanhRND. However, using $k$-medoids for mini-batch diversification generates fewer diverse activities than the diversification-free methods, even when not modifying the rewards.

\subsubsection{Glycogen Synthase Kinase 3 Beta (GSK3$\beta$)}
\Cref{fig:gsk3b_circles} displays the cumulative number of diverse actives per generative step on the GSK3$\beta$-based reward function. We observe that utilizing mini-batch diversification via DPP or MaxMin generates significantly more diverse active after a few hundred generative steps. We see this behaviour no matter if we modify the extrinsic reward or not, meaning that mini-batch diversification can successfully be used as an exploration technique to overcome mode collapse and lead to diverse behaviors. Moreover, we notice that, after at most 4000 generative steps, MaxMin yields substantially more diverse actives than the other methods.  Also, we note that, similar to the experiments on the DRD2-based reward functions, using $k$-medoids yields a substantially lower number of diverse actives than the other methods, including diversification-free methods.

\subsubsection{c-Jun N-terminal Kinases-3 (JNK3)}
\Cref{fig:jnk3_circles} shows the cumulative number of diverse actives per generative step on the JNK3-based reward function. Firstly, we observe a high standard deviation among all experiments, compared to the other reward functions. This is likely since the JNK3 oracle is more difficult to optimize than the other oracles, and therefore does not have a large margin to the activity threshold of 0.5 for diverse actives. Similar trends in terms of diversity have been observed by previous work \cite{svensson2024diversityawarereinforcementlearningnovo}. Most approaches using mini-batch diversification keep improving over a large number of generative steps, while the diversification-free experiments generally show a substantially lower number of average diverse actives. For no extrinsic reward modification (see left plot in \cref{fig:jnk3_circles}), MaxMin generates the highest average number of diverse actives, while DPP has lower variance but yields fewer diverse actives. When using the IMS or TanhRND strategy to modify the reward (see middle and right plot in \cref{fig:jnk3_circles}), MaxMin also yields the highest average number of diverse actives, but the runs overlap with DPP since both have high variance. For the experiments using TanhRND (see right plot in \cref{fig:jnk3_circles}), all MaxMin configurations display a larger increase in the average number of diverse actives over time. On this reward function, $k$-medoids can generate more diverse actives than the diversification-free method when not modifying the (extrinsic) reward, while these two methods display similar performance when modifying the reward.  

\subsection{Diverse Mini-Batch Selection Enhances Reference-Based Diversity}
\begin{figure}[t]
     \centering
     \begin{subfigure}[b]{0.47\textwidth}
         \centering
         \includegraphics[width=\textwidth]{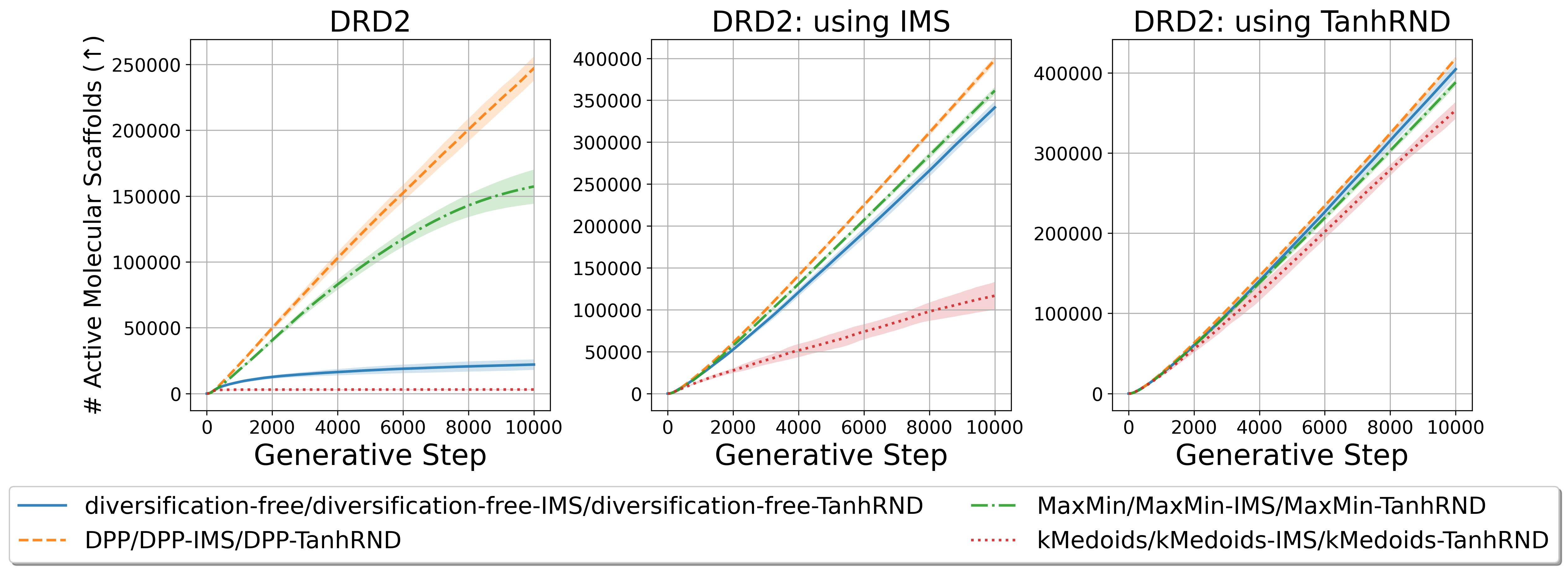}
         \caption{DRD2}
         \label{fig:drd2_scaffolds}
     \end{subfigure}
     \hfill
     \begin{subfigure}[b]{0.47\textwidth}
         \centering
         \includegraphics[width=\textwidth]{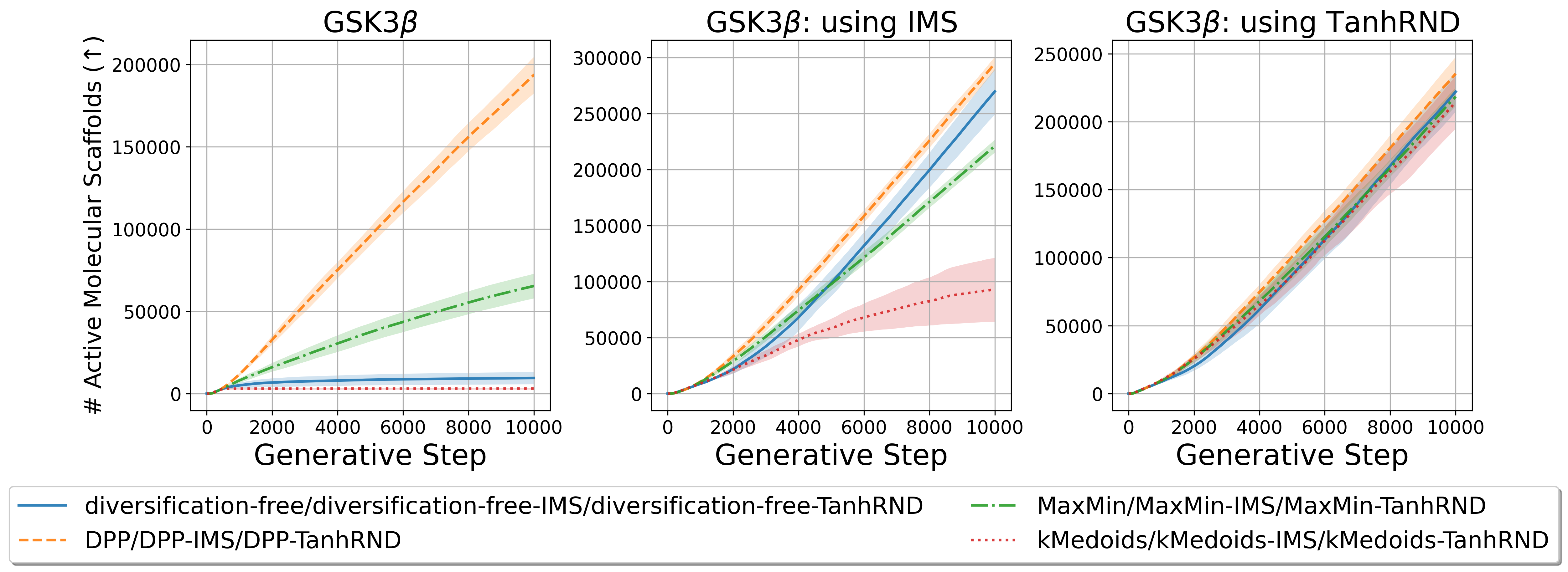}
         \caption{GSK3$\beta$}
         \label{fig:gsk3b_scaffolds}
     \end{subfigure}
     \hfill
     \begin{subfigure}[b]{0.47\textwidth}
         \centering
         \includegraphics[width=\textwidth]{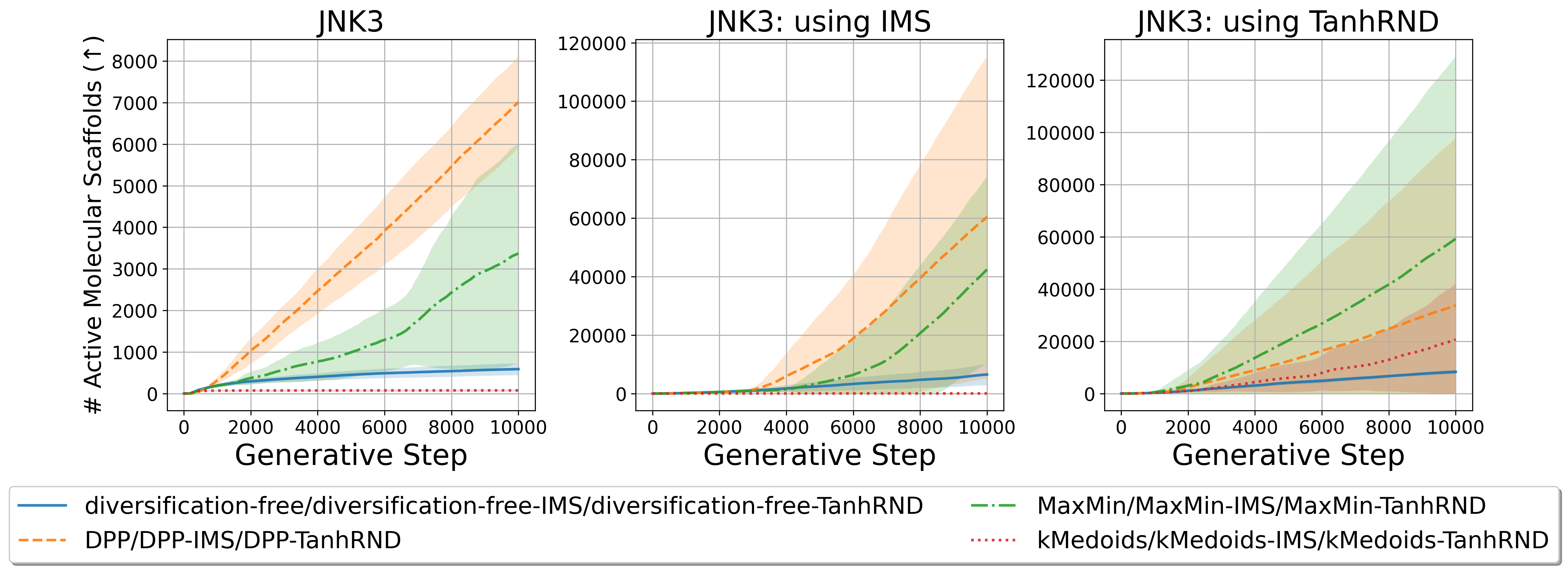}
         \caption{JNK3}
         \label{fig:jnk3_scaffolds}
     \end{subfigure}
        \caption{Total number of molecular scaffolds after $g$ generative steps evaluated on reward functions based on the DRD2, GSK3$\beta$, or JNK3 predictive model. The average across 10 independent runs per generative step is plotted over \num{10000} generative steps, where the shaded area shows standard deviations among the independent runs. }
        \label{fig:scaffolds}
\end{figure}
To obtain a more comprehensive evaluation of the diversity, we also investigate reference-based diversity \cite{hu2024hamiltonian}. In particular, we consider the number of unique molecular scaffolds, also named Bemis-Murcko scaffolds \cite{bemis1996properties}, computed by RDKit \cite{landrum2006rdkit}. We are only interested in the diversity of molecules suitable for our target and, therefore, only consider scaffolds of active molecules with both an oracle score and QED above 0.5. \Cref{fig:scaffolds} shows the cumulative number of unique active molecular scaffolds per generative step for the DRD2-, GSK3$\beta$- and JNK3-based reward functions. The lines and shaded area display the mean and standard deviation, respectively, across 10 independent reruns for each configuration.

\subsubsection{Dopamine Receptor D2 (DRD2)}
\Cref{fig:drd2_scaffolds} displays the cumulative number of active molecular scaffolds, per generative step, evaluated on the DRD2-based reward function. When not modifying the extrinsic reward (see left plot in \cref{fig:drd2_scaffolds}), using mini-batch diversification via DPP or MaxMin leads to substantially more scaffolds, compared to the diversification-free method, after less than 750 generative steps. In particular, our experiments demonstrate that DPP generates most scaffolds on average. When utilizing the identical molecular scaffold (IMS) filter \cite{blaschke2020memory} for modifying the extrinsic reward (see middle plot in \cref{fig:drd2_scaffolds}), we observe that DPP generates more molecular scaffolds compared to the other methods. For the TanhRND technique (see right plot in \cref{fig:drd2_scaffolds}), the diversification-free, MaxMin and DPP methods show similar diversity in terms of molecular scaffolds and perform on par with the best methods when using IMS (see middle plot in \cref{fig:drd2_scaffolds}).
In terms of molecular scaffolds, it is clear that the scaffold-based similarity that mini-batch diversification provides can be important, especially in combination with no or less effective exploration techniques. However, across all experiments, it is clear that $k$-medoids generates the least amount of scaffolds, and it is therefore important to choose an appropriate method for mini-batch diversification.

\subsubsection{Glycogen Synthase Kinase 3 Beta (GSK3$\beta$)}
\Cref{fig:gsk3b_scaffolds} displays the cumulative number of molecular scaffolds for the evaluation on the GSK3$\beta$-based reward function. Without any modification of the extrinsic reward (see left plot in \cref{fig:gsk3b_scaffolds}), we observe that mini-batch diversification via DPP or MaxMin yields significantly more scaffolds compared to the diversification-free method (blue line). The DPP effectively generates more molecular scaffolds, while MaxMin is less effective. For reward modification (see the middle and right plots in \cref{fig:gsk3b_scaffolds}), we observe that using mini-batch diversification via DPP generates more scaffolds on average and has lower variance. However, the difference in effectiveness of using DPP is reduced in terms of diverse actives, but DPP can still consistently improve diversity. For MaxMin, which consistently generates the largest number of diverse actives (see \cref{fig:gsk3b_circles}), we observe a lower number of scaffolds. Thus, when using the MaxMin algorithm to impose mini-batch diversity, we see that high distance-based diversity does not directly result in high reference-based diversity, and vice versa. When using mini-batch diversification via $k$-medoids, it generates significantly fewer scaffolds, except when using TanhRND, where it performs on par with the other methods. 

\subsubsection{c-Jun N-terminal Kinases-3 (JNK3)}
\Cref{fig:jnk3_scaffolds} displays the scaffold diversity for the evaluation on the JNK3-based reward function. When not modifying the extrinsic reward (see left plot in \cref{fig:jnk3_scaffolds}), all DPP-based methods are more effective after around 2000 generative steps. For DPP, we observe the largest average number of molecular scaffolds and notice a more consistent exploration, since the rate of diverse solutions is higher. The MaxMin algorithm does not display the same consistent improvement in the number of scaffolds. When modifying the extrinsic reward (see middle and right plots in \cref{fig:jnk3_scaffolds}), both DPP and MaxMin obtain a higher average number of scaffolds, but they also display a high variability and are therefore not always more effective. This is likely because the agent is not able to effectively optimize the reward (see \cref{fig:jnk3_reward}). In general, as depicted in \cref{fig:jnk3_reward}, the JNK3-based reward is more difficult to optimize for the RL agent. Thus, we notice that the robustness of our proposed mini-batch diversification depends on how well the agent can optimize the given task. This is expected since the mini-batch selection depends on the given larger set $\mathcal{B}$ and, therefore, has limited capabilities to enhance the diversity if the RL agent itself cannot find sufficient solutions.

\section{Conclusions}
In this work, we present an easily applicable framework for enhancing mini-batch diversity in reinforcement learning algorithms. The framework seeks to tackle the problem of efficient exploration when it is costly to evaluate a reward function. In this paper, we apply our framework to \textit{de novo} drug design, but the framework is problem-agnostic. We believe that the proposed framework can also be beneficial in other applications in reinforcement learning, where efficient exploration and diverse behaviors are crucial. To solve the problem of mini-batch diversification in RL, we study the use of determinantal point processes (DPPs) \cite{kulesza2012determinantal}, the MaxMin algorithm \cite{ashton2002identification} and $k$-medoids clustering \cite{rdusseeun1987clustering} for the diversification process.
In this way, we seek to summarize a larger set of molecules by selecting a smaller mini-batch of diverse molecules to evaluate, requiring fewer evaluations. DPP samples a diverse mini-batch given a kernel matrix, while the MaxMin algorithm and $k$-medoids clustering aim to find the maximum coverage of the larger set with respect to dissimilarities between molecules. We argue that this enhances the exploration by focusing on promising, more diverse molecules, while keeping the rewards high.  
We observe that our proposed framework for mini-batch diversification can substantially improve the diversity of \textit{de novo} drug design, especially when combined with a domain-specific modification of the extrinsic reward, such as TanhRND \cite{svensson2024diversityawarereinforcementlearningnovo}. We demonstrate that DPP-based mini-batch diversification enhances both distance- and reference-based diversity, while the MaxMin algorithm primarily improves distance-based diversity. Therefore, we propose to use DPP for the diversification process, since it also allows for a more adaptable kernel matrix, e.g., by incorporating quality terms, and a natural way to introduce randomness in the diversification process. Moreover, we notice that if the agent alone provides sufficient solutions, our framework can substantially enhance the diversity of the generated solutions.
Our experiments indicate that using diverse mini-batches in reinforcement learning improves exploration and provides a basis for the effectiveness of this approach.

\section*{Acknowledgements}
This work was partially supported by the Wallenberg AI, Autonomous Systems and Software Program (WASP) funded by the Knut and Alice Wallenberg Foundation. The experimental evaluation was enabled by resources provided by the National Academic Infrastructure for Supercomputing in Sweden (NAISS), partially funded by the Swedish Research Council through grant agreement no. 2022-06725. We thank

\bigskip

\bibliography{main}
\clearpage

\input{appendix}

\end{document}

%% file: dpp_rl.tikz
\begin{tikzpicture}[style1/.style={rectangle, rounded corners,draw=black,thick,minimum size=10mm,font=\Large}, style2/.style={circle,thick, draw,minimum size=7mm,font=\Large}]
    \node[draw,style1] (r) at (0,0) {RL};
    \node[draw, style1,minimum size=20mm, right = of r] (set) {};
    \node[draw, rectangle, fill=blue!50] (r1) at (1.9,-0.7) {};
    \node[draw, rectangle, fill=blue!50] (r2) at (1.9,-0.3) {};
    \node[draw, rectangle, fill=blue!50] (r3) at (2.4,-0.5) {};
    \node[draw, regular polygon, fill=red!50] (t1) at (3.2,0.6) {};
    \node[draw, regular polygon, fill=red!50] (t2) at (3,0.2) {};
    \node[draw, circle, fill=green!50] (c1) at (1.8,0.7) {};
    \draw[-latex, thick] (r.east) -- (set.west);
    \node[rectangle, minimum size=20mm,font=\Large, right = of set] (dpp) {}; 
    \draw [decorate,decoration={brace,amplitude=4pt},xshift=0pt,yshift=0pt, thick] (4.8,-0.5) -- (4.8,0.5) node [midway,right,xshift=.1cm] {};
    \node[draw,rectangle,fill=blue!50] at (5,0) {};
    \node[rectangle,font=\huge] at (5.2,-0.1) {,};
    \node[draw,regular polygon,fill=red!50] at (5.5,0) {};
    \node[rectangle,font=\huge] at (5.8,-0.1) {,};
    \node[draw,circle,fill=green!50] at (6.1,0) {};
    \draw [decorate,decoration={brace,amplitude=4pt},xshift=0pt,yshift=0pt, thick] (6.3,0.5) -- (6.3,-0.5) node [midway,right,xshift=.1cm] {};
    \draw[-latex, thick] (set.east) -- (dpp.west);
    \node[draw, style1, rounded corners, right = of dpp] (s) {Evaluation};
    \draw[-latex,thick] (dpp) -- (s);
    \draw[thick] let \p1 = (s), \p2 = (r)in 
    [thick] (s.south) -- ($(\x1,\y1) - (0,1.5)$)
    [thick] ($(\x1,\y1) - (0,1.5)$) -- ($(\x2,\y2) - (0,1.5)$)
    [-latex,thick] ($(\x2,\y2) - (0,1.5)$) -- (r.south);
\end{tikzpicture}

%% file: appendix.tex
\appendix

\section{Kernel Matrix for DPP}
\label{app:kernel_matrix_dpp}
To obtain a diverse mini-batch, we perform exact sampling from a $k$-DPP using the Gram-Schmidt sampler implemented in DPPy \cite{GPBV19}. Performing exact sampling from the $k$-DPP typically requires an eigendecomposition of its kernel \cite{kulesza2011k}, typically requiring $\mathcal{O}(N^3)$ time. Given a decomposition, drawing a sample typically takes $\mathcal{O}(NK^3)$ time overall \cite{kulesza2012determinantal}. For more details, we refer to \cite{derezinski2019exact,calandriello2020sampling} for more efficient exact sampling procedures, \cite{li2016efficient,grosse2024greedy} for approximative methods and \cite{anari2016monte,rezaei2019polynomial} for Markov-Chain-Monte-Carlo (MCMC) procedures. 

To perform sampling from a $k$-DPP, a kernel matrix $L$ needs to be constructed at each generative step.
We explore two different approaches to measure the similarity between molecules, resulting in two base kernel matrices that incorporate varying levels of information. The first base kernel matrix is constructed by the Tanimoto similarity between the corresponding 2048-bit Morgan fingerprints (with radius 2 using RDKit \cite{landrum2006rdkit}) of the generated SMILES. We denote this base matrix by $L_T$. To incorporate more scaffold-based information, we also create a base kernel matrix by computing the Dice coefficients \cite{dice1945measures,sorensen1948method} between the scaffolds' atom pair fingerprints \cite{carhart1985atom}. We denote this base kernel matrix by $L_D$.

\begin{figure}[ht]
     \centering
     \begin{subfigure}[b]{0.47\textwidth}
         \centering
         \includegraphics[width=\textwidth]{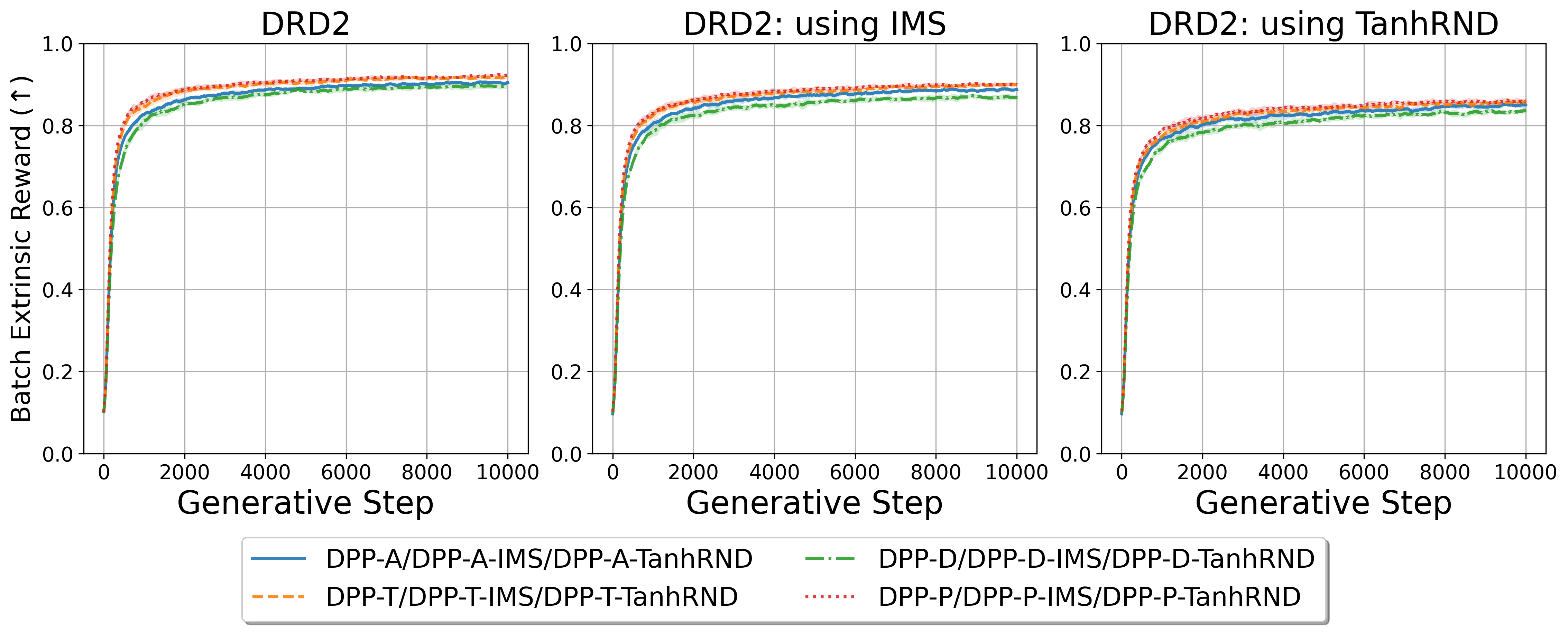}
         \caption{DRD2}
         \label{fig:drd2_reward_dpp}
     \end{subfigure}
     \hfill
     \begin{subfigure}[b]{0.47\textwidth}
         \centering
         \includegraphics[width=\textwidth]{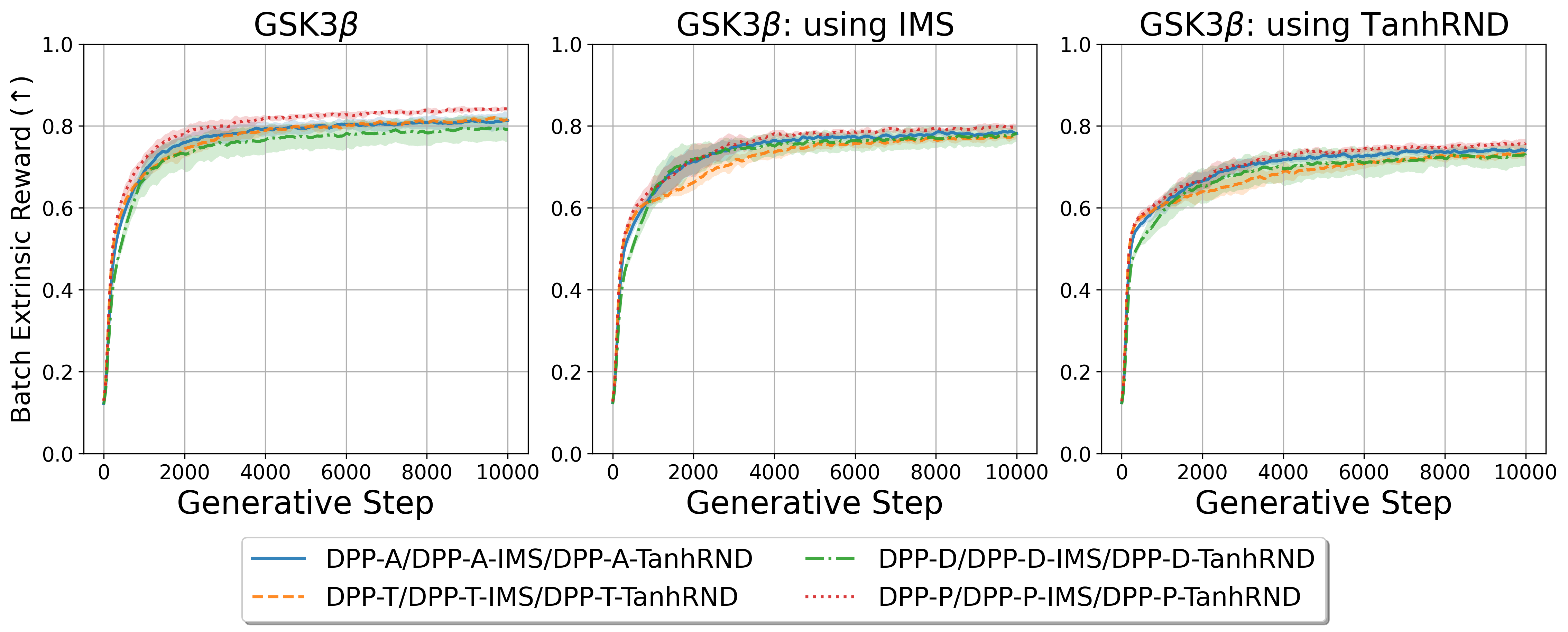}
         \caption{GSK3$\beta$}
         \label{fig:gsk3b_reward_dpp}
     \end{subfigure}
     \hfill
     \begin{subfigure}[b]{0.47\textwidth}
         \centering
         \includegraphics[width=\textwidth]{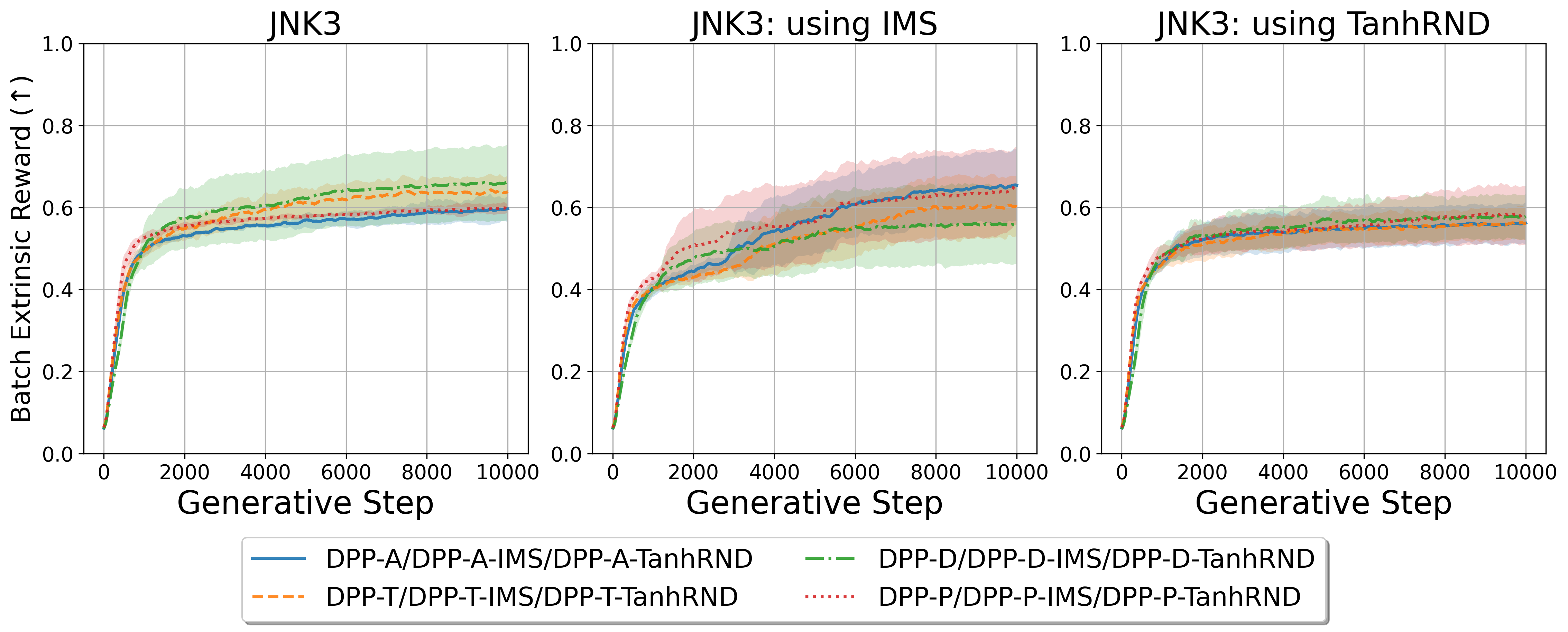}
         \caption{JNK3}
         \label{fig:jnk3_reward_dpp}
     \end{subfigure}
        \caption{Average extrinsic rewards per generative step across the mini-batch of SMILES evaluated on the DRD2-, GSK3$\beta$-, or JNK3-based reward functions. For clarity of presentation, we display the moving averages with a window size of 101. }
        \label{fig:reward_dpp}
\end{figure}

\begin{figure}[t]
     \centering
     \begin{subfigure}[b]{0.47\textwidth}
         \centering
         \includegraphics[width=\textwidth]{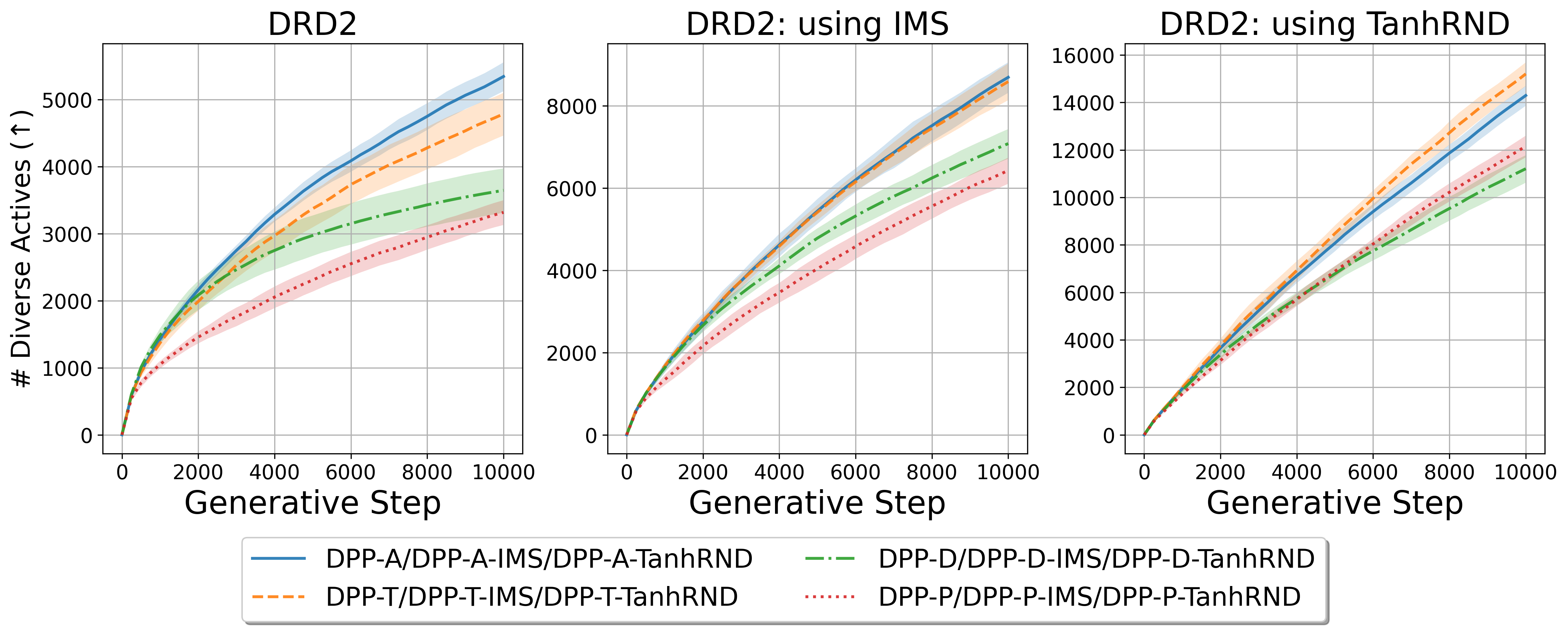}
         \caption{DRD2}
         \label{fig:drd2_circles_dpp}
     \end{subfigure}
     \hfill
     \begin{subfigure}[b]{0.47\textwidth}
         \centering
         \includegraphics[width=\textwidth]{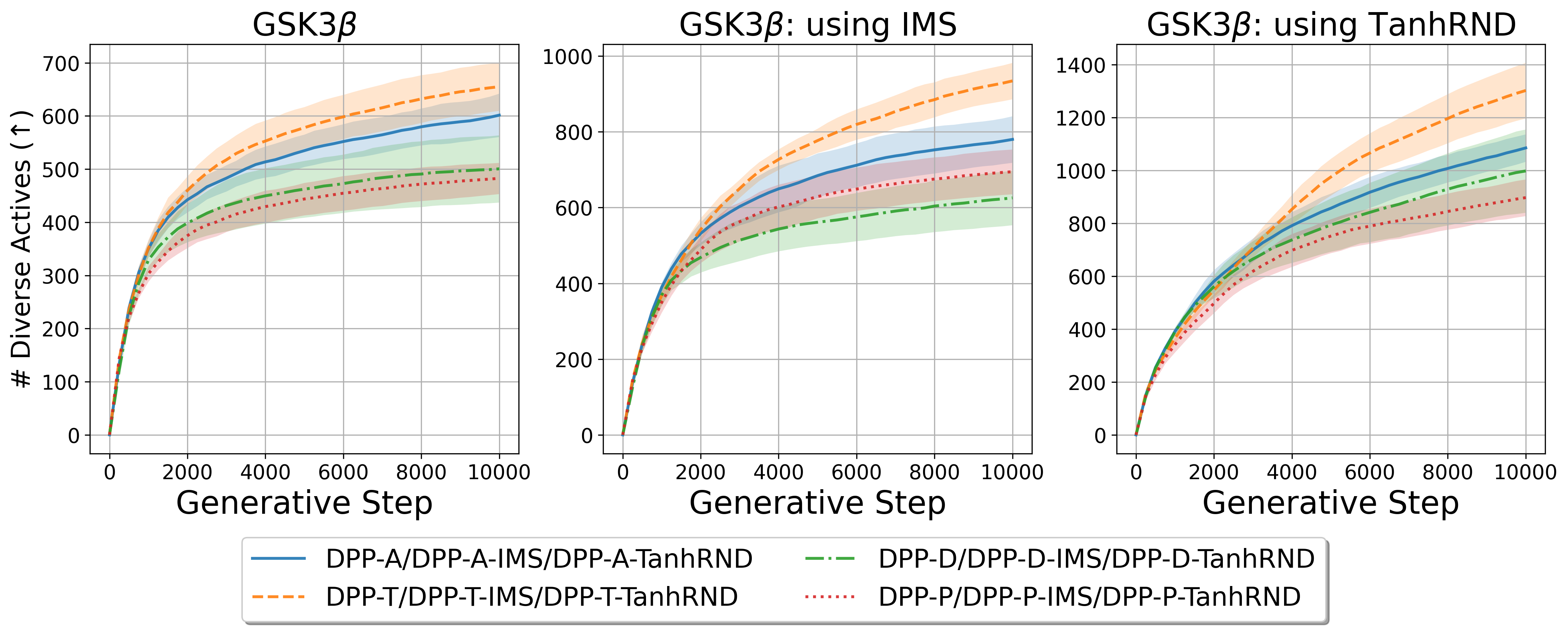}
         \caption{GSK3$\beta$}
         \label{fig:gsk3b_circles_dpp}
     \end{subfigure}
     \hfill
     \begin{subfigure}[b]{0.47\textwidth}
         \centering
         \includegraphics[width=\textwidth]{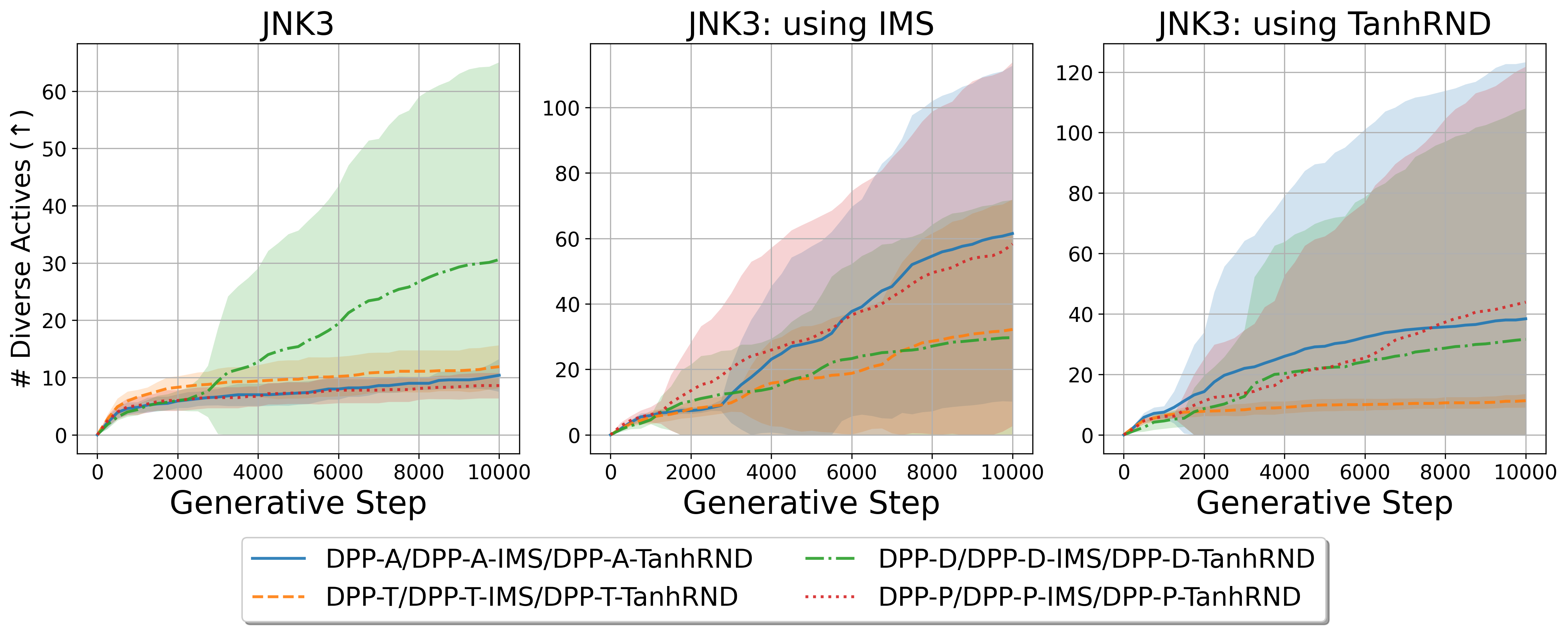}
         \caption{JNK3}
         \label{fig:jnk3_circles_dpp}
     \end{subfigure}
        \caption{Total number of diverse activities after $g$ generative steps evaluated on reward functions based on the DRD2, GSK3$\beta$, or JNK3 predictive model. The total number of diverse actives is plotted for every 250th generative step.}
        \label{fig:circles_dpp}
\end{figure}

\begin{figure}[t]
     \centering
     \begin{subfigure}[b]{0.47\textwidth}
         \centering
         \includegraphics[width=\textwidth]{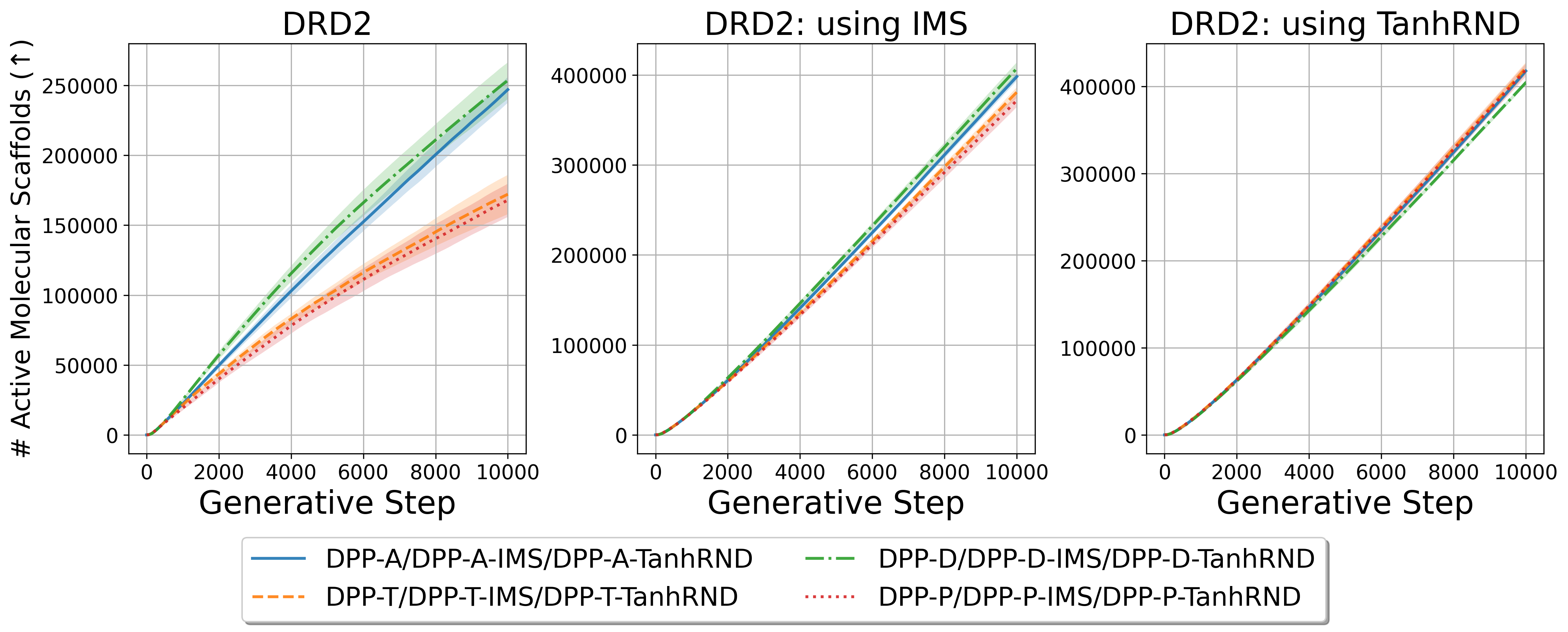}
         \caption{DRD2}
         \label{fig:drd2_scaffolds_dpp}
     \end{subfigure}
     \hfill
     \begin{subfigure}[b]{0.47\textwidth}
         \centering
         \includegraphics[width=\textwidth]{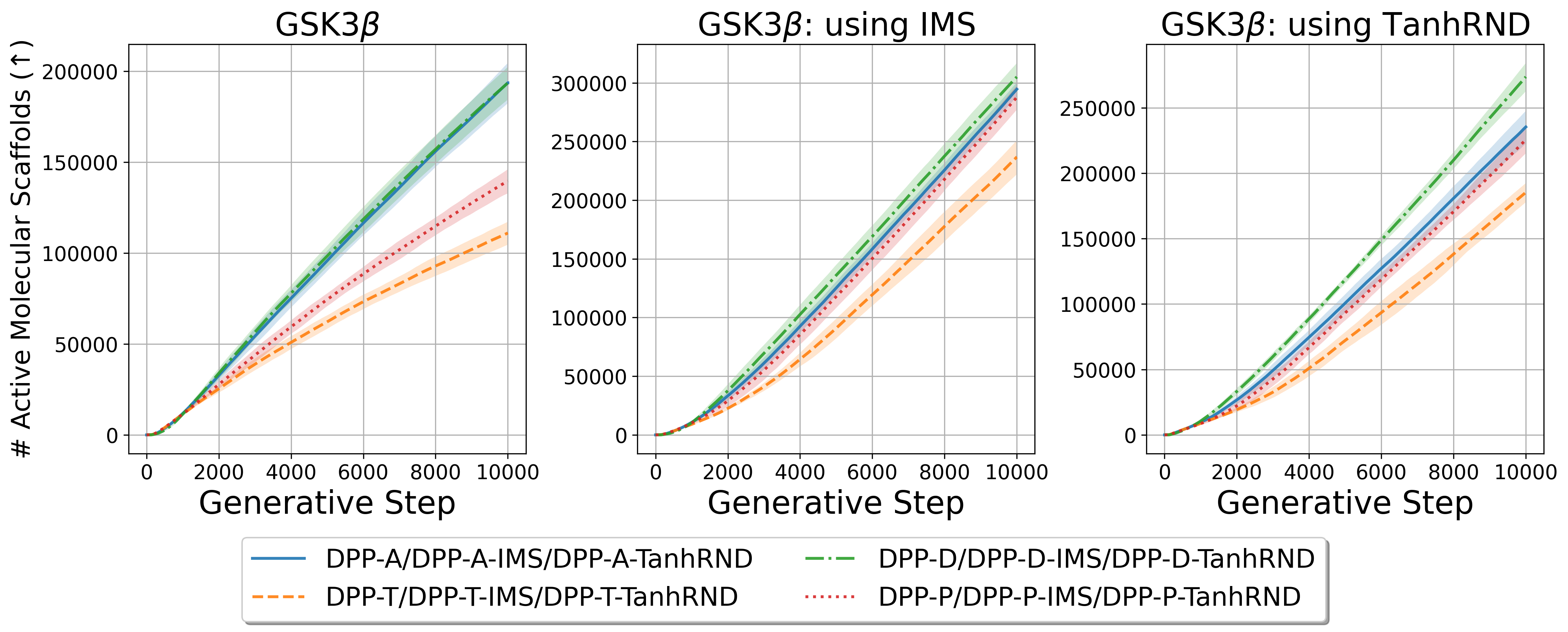}
         \caption{GSK3$\beta$}
         \label{fig:gsk3b_scaffolds_dpp}
     \end{subfigure}
     \hfill
     \begin{subfigure}[b]{0.47\textwidth}
         \centering
         \includegraphics[width=\textwidth]{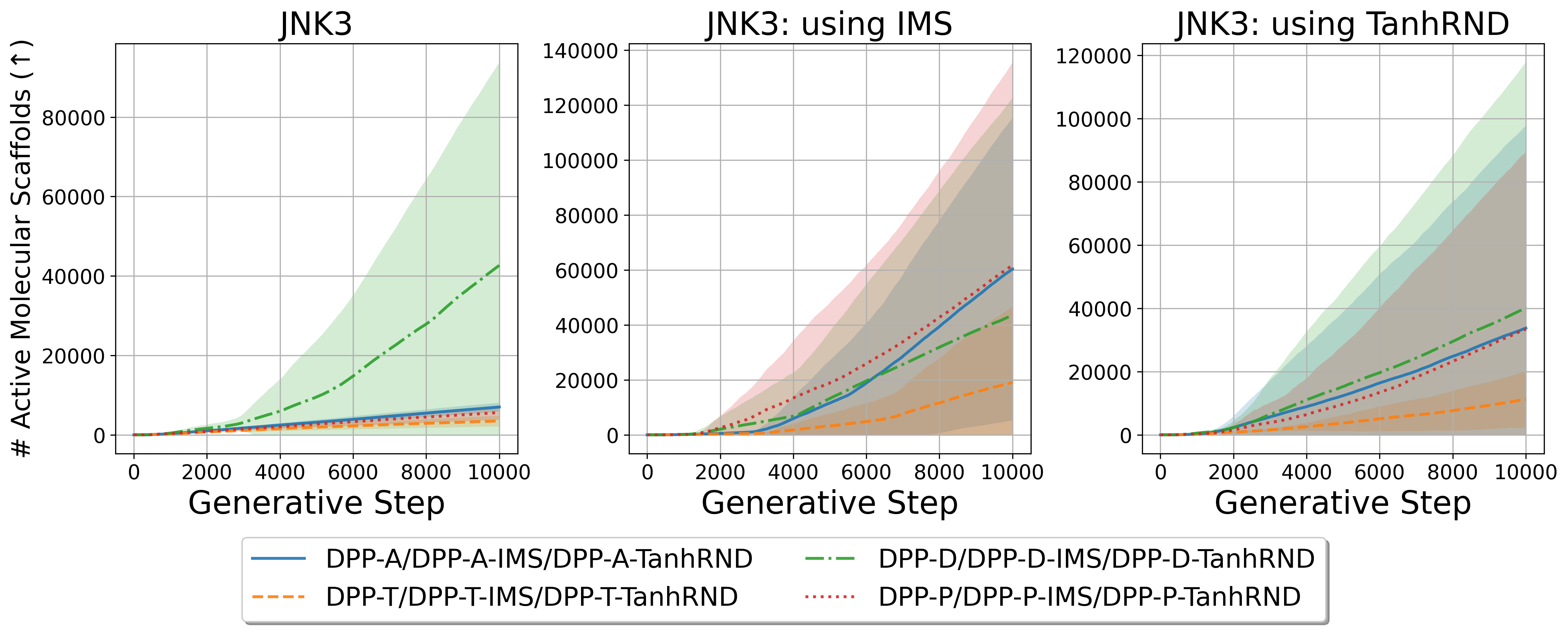}
         \caption{JNK3}
         \label{fig:jnk3_scaffolds_dpp}
     \end{subfigure}
        \caption{Total number of molecular scaffolds after $g$ generative steps evaluated on reward functions based on the DRD2, GSK3$\beta$, or JNK3 predictive model.}
        \label{fig:scaffolds_dpp}
\end{figure}

We investigate four combinations of $L_T$ and $L_D$ to create the kernel matrix $L$ used for sampling from a $k$-DPP (see \cref{tab:L_matrix}). We obtain the first variant by element-wise summation of $L_T$ and $L_D$, which we denote by DPP-A. Note that taking an element-wise arithmetic mean instead, i.e., multiplying a constant term $1/2$ with all items in $L$, does not change the probabilities and, therefore, would make no difference in practice for sampling. The second variant is obtained by only using $L_T$, which we denote by DPP-T. The third variant is obtained by the element-wise product of the two matrices, which we denote by DPP-P. The last variant is obtained only using $L_D$ and is denoted by DPP-D. This results in four different configurations of DPP. 
\begin{table}
    \setlength{\tabcolsep}{1mm}
    \centering
    \begin{tabular}{ccccc} \toprule
          DPP-A & DPP-T & DPP-P  & DPP-D \\ \midrule
           $L_T + L_D$ & $L_T$ & $L_T \odot L_D$ & $L_D$\\ \bottomrule
    \end{tabular}
    \caption{Different kernel matrix $L$ configurations. $L_T$ consists of Tanimoto similarities between Morgan fingerprints and $L_D$ consists of Dice similarities between atom-pair fingerprints. $\odot$ denotes element-wise multiplication. }
    \label{tab:L_matrix}
\end{table}
For each kernel matrix in \cref{tab:L_matrix} for DPP, we study how it affects the quality and diversity on the different reward functions. We investigate mini-batch diversification in combination with different techniques to modify the reward function (for enhancing exploration and diversity). \Cref{fig:reward_dpp} displays the average extrinsic reward and standard deviation per generative step on the DRD2-, GSK3$\beta$-, or JNK3-based reward functions. For clarity of presentation, we display the moving averages with a window size of 101. Each line shows the average, while the shaded area shows the standard deviation. For all different configurations of DPP, we observe similar trends in terms of extrinsic rewards. \Cref{fig:circles_dpp} displays the total number of diverse activities up to the current generative steps. The total number of diverse actives is plotted for every 250th generative step. For the DRD2-based reward functions, both DPP-A and DPP-T generate among the largest number of diverse actives. For the GSK3$\beta$-based reward function, DPP-T often generates the largest number of diverse actives, while DPP is the second-best configuration. For the JNK3-based reward function, the variability of all methods is high when they can generate more than around 10 diverse actives. We observe that DPP-A often displays the largest number of diverse actives when using the IMS and TanhRND technique to modify the reward. \Cref{fig:scaffolds_dpp} shows the total number of molecular scaffolds up to the current generative step. For the DRD2-based reward function, DPP-A consistently generates the largest number of scaffolds. On GSK3$\beta$, DPP-D generates the largest number of scaffolds, while DPP-A is the second-best method. There is a high variability for the JNK3 reward. DPP-A displays a large average number of scaffolds. Overall, we observe that DPP-A consistently displays a good balance between the different diversity metrics. Therefore, we use this method in the main paper to represent mini-batch diversification via DPP sampling. 

\section{Kernel Matrix for Maximum Coverage}
\label{app:kernel_matrix_max_cov}
We also investigate three different dissimilarity functions for the MaxMin algorithm and $k$-medoids clustering. We also refer to kernel matrices for the MaxMin algorithm and $k$-medoids clustering. Thus, we explore the following configurations of the MaxMin algorithm and $k$-medoids clustering: (1) ``MaxMin-T''/``kMedoids-T'' using the Tanimoto similarity between the Morgan fingerprints described above, which corresponds to using kernel matrix $L_T$; (2) ``MaxMin-D''/``kMedoids-D'' using the Dice similarity with the atom pair fingerprints described above, which corresponds to $L_D$; (3) ``MaxMin-A''/"kMedoids-A" using the average Tanimoto and Dice similarities, which corresponds to $\frac{L_D + L_T}{2}$. This results in 3 different configurations for the MaxMin algorithm and $k$-medoids clustering. We also denote these dissimilarity functions as kernel matrices.
For each dissimilarity function for the MaxMin algorithm and $k$-medoids clustering, we study how it affects the quality and diversity on the different reward functions. We investigate mini-batch diversification in combination with different techniques to modify the reward function (for enhancing exploration and diversity).

\subsection{MaxMin Algorithm}
\begin{figure}[ht]
     \centering
     \begin{subfigure}[b]{0.47\textwidth}
         \centering
         \includegraphics[width=\textwidth]{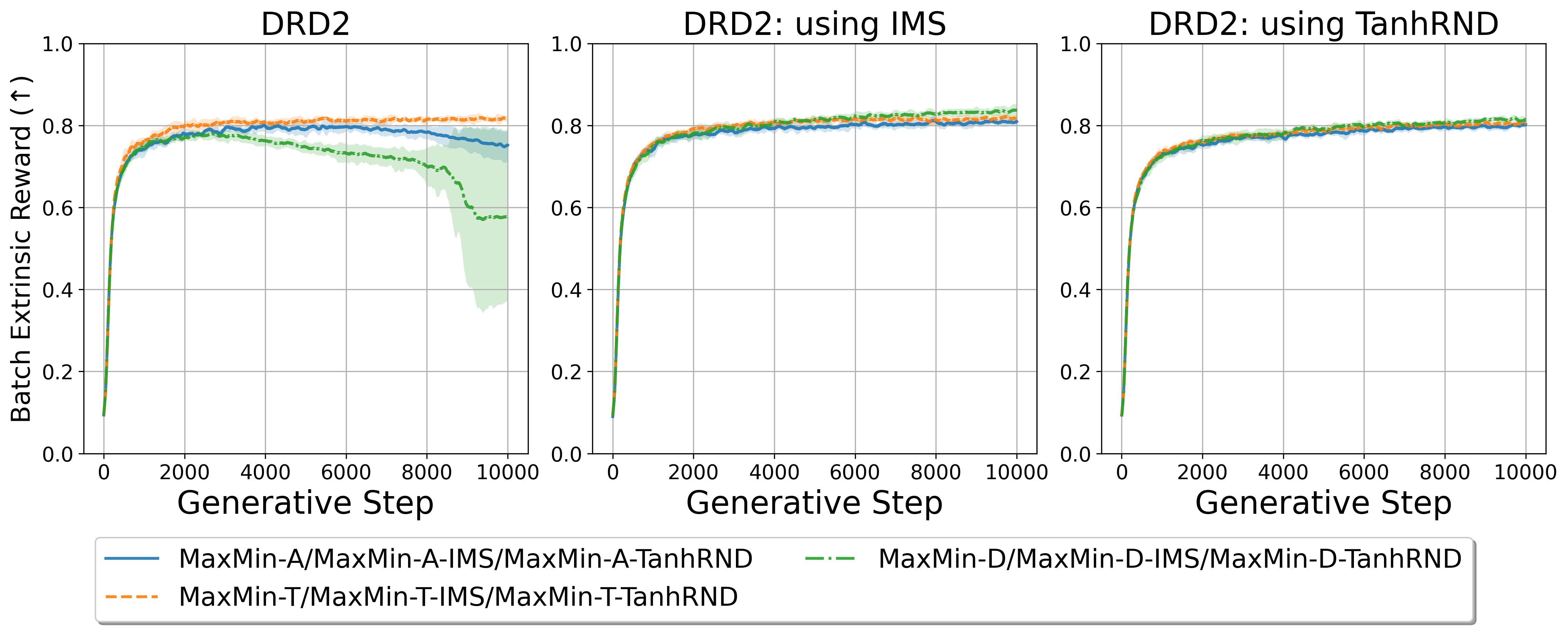}
         \caption{DRD2}
         \label{fig:drd2_reward_maxmin}
     \end{subfigure}
     \hfill
     \begin{subfigure}[b]{0.47\textwidth}
         \centering
         \includegraphics[width=\textwidth]{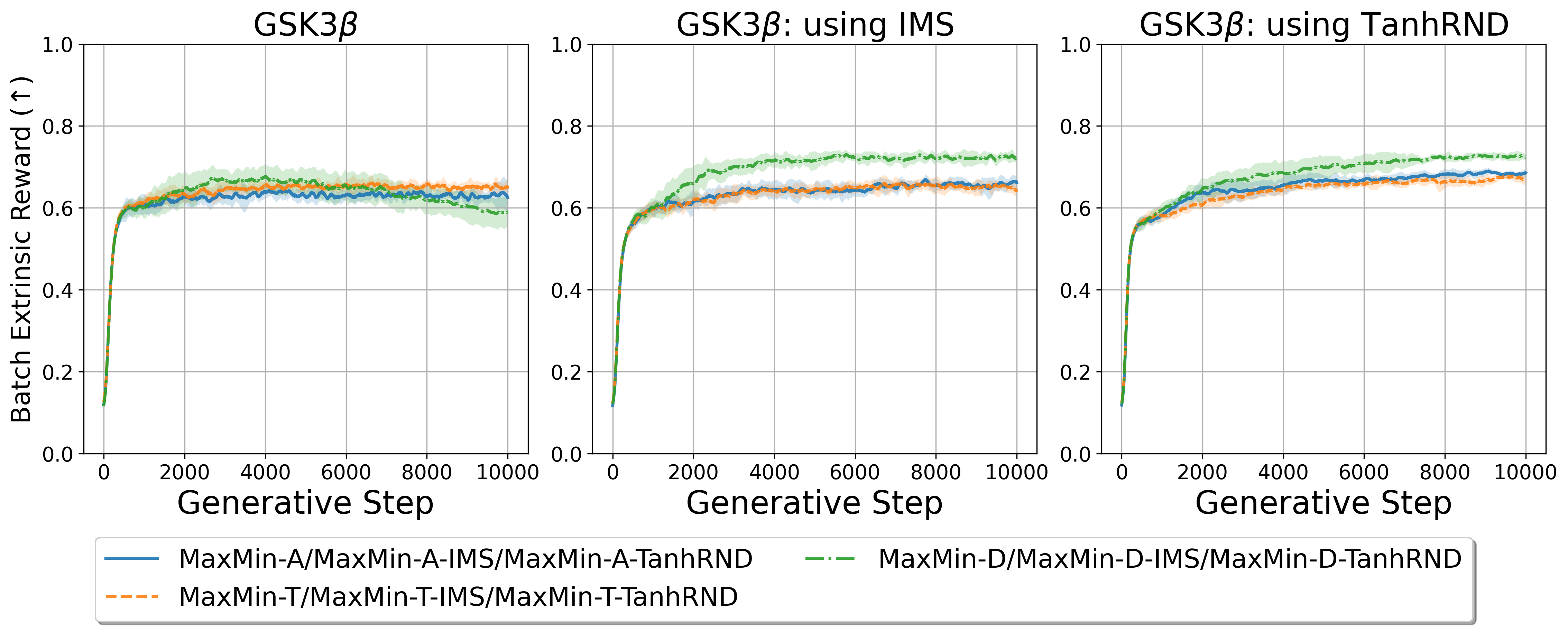}
         \caption{GSK3$\beta$}
         \label{fig:gsk3b_reward_maxmin}
     \end{subfigure}
     \hfill
     \begin{subfigure}[b]{0.47\textwidth}
         \centering
         \includegraphics[width=\textwidth]{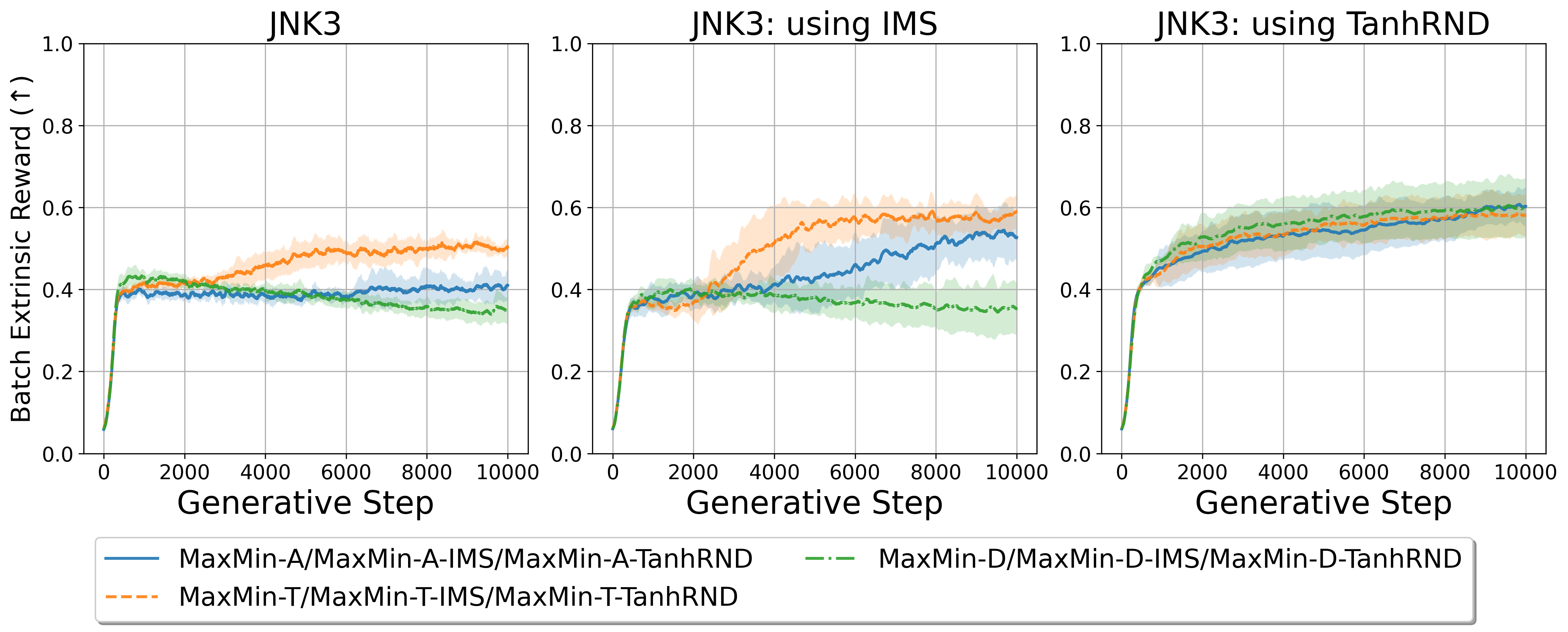}
         \caption{JNK3}
         \label{fig:jnk3_reward_maxmin}
     \end{subfigure}
        \caption{Average extrinsic rewards per generative step across the mini-batch of SMILES evaluated on the DRD2-, GSK3$\beta$-, or JNK3-based reward functions. For clarity of presentation, we display the moving averages with a window size of 101. }
        \label{fig:reward_maxmin}
\end{figure}

\begin{figure}[t]
     \centering
     \begin{subfigure}[b]{0.47\textwidth}
         \centering
         \includegraphics[width=\textwidth]{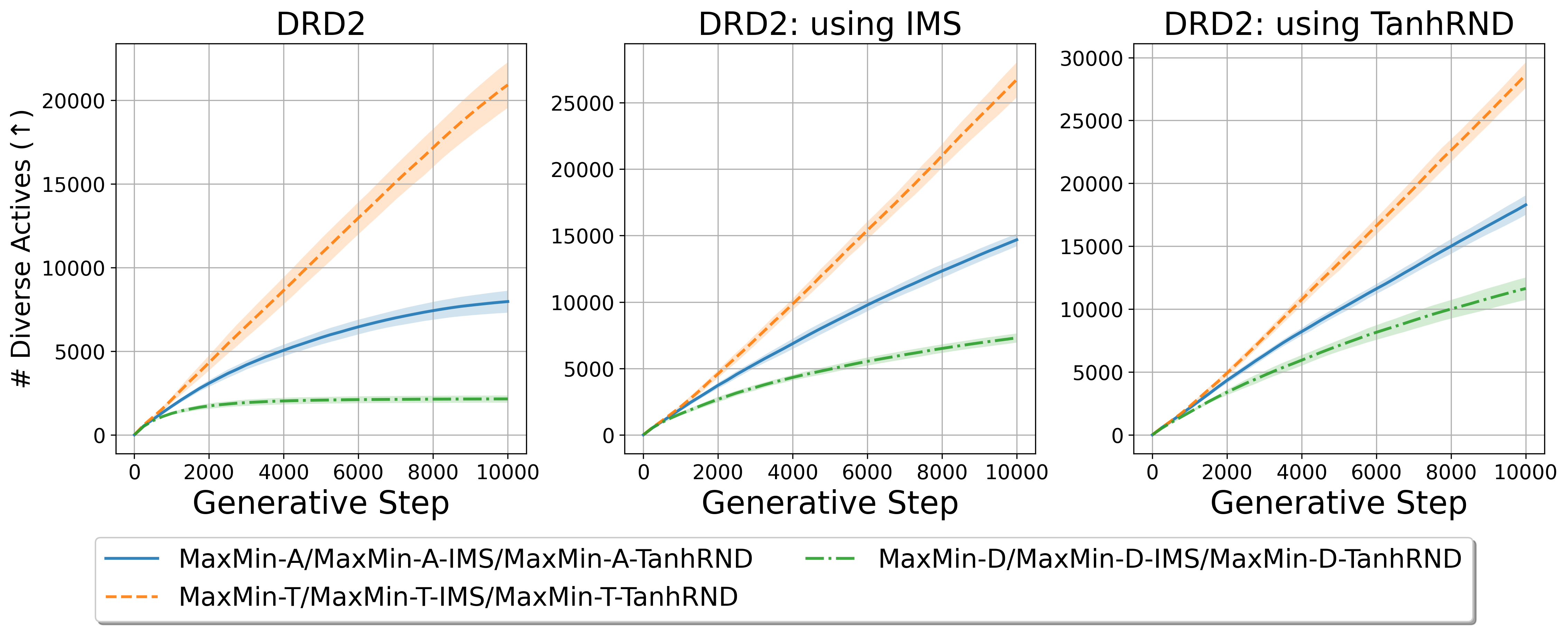}
         \caption{DRD2}
         \label{fig:drd2_circles_maxmin}
     \end{subfigure}
     \hfill
     \begin{subfigure}[b]{0.47\textwidth}
         \centering
         \includegraphics[width=\textwidth]{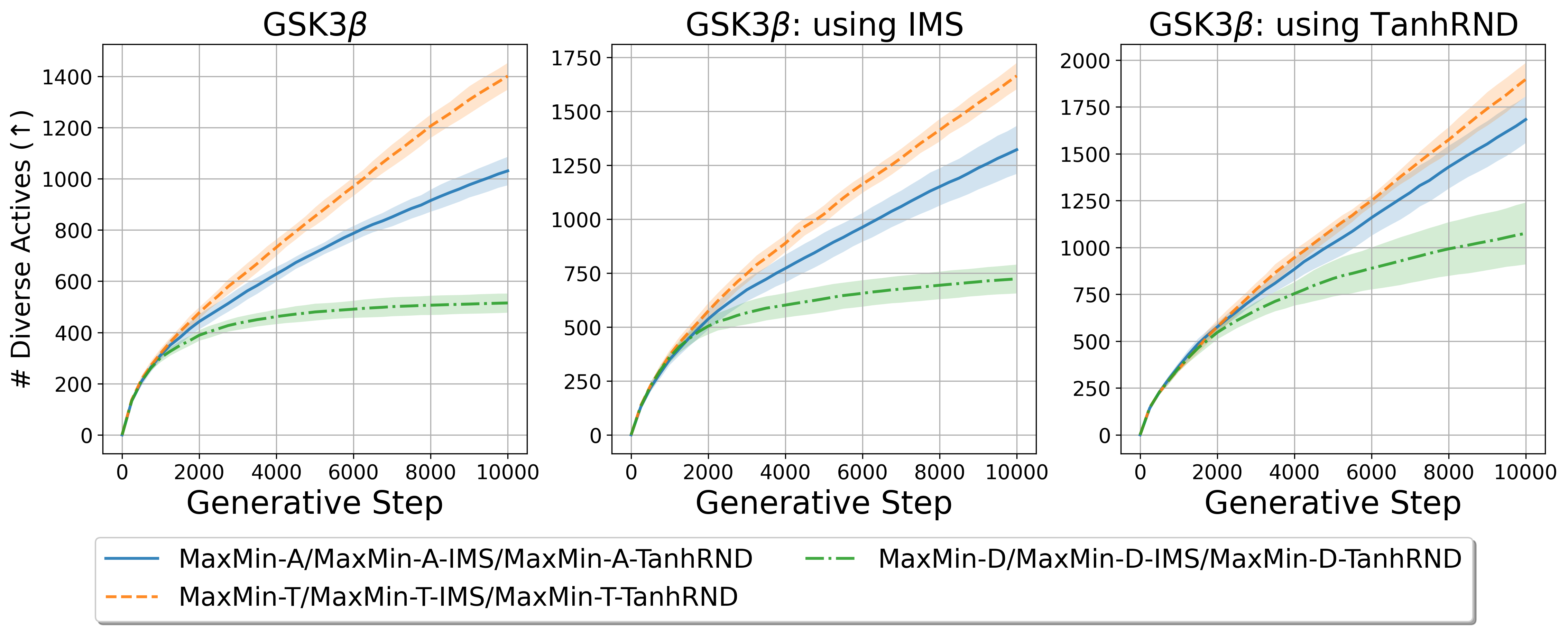}
         \caption{GSK3$\beta$}
         \label{fig:gsk3b_circles_maxmin}
     \end{subfigure}
     \hfill
     \begin{subfigure}[b]{0.47\textwidth}
         \centering
         \includegraphics[width=\textwidth]{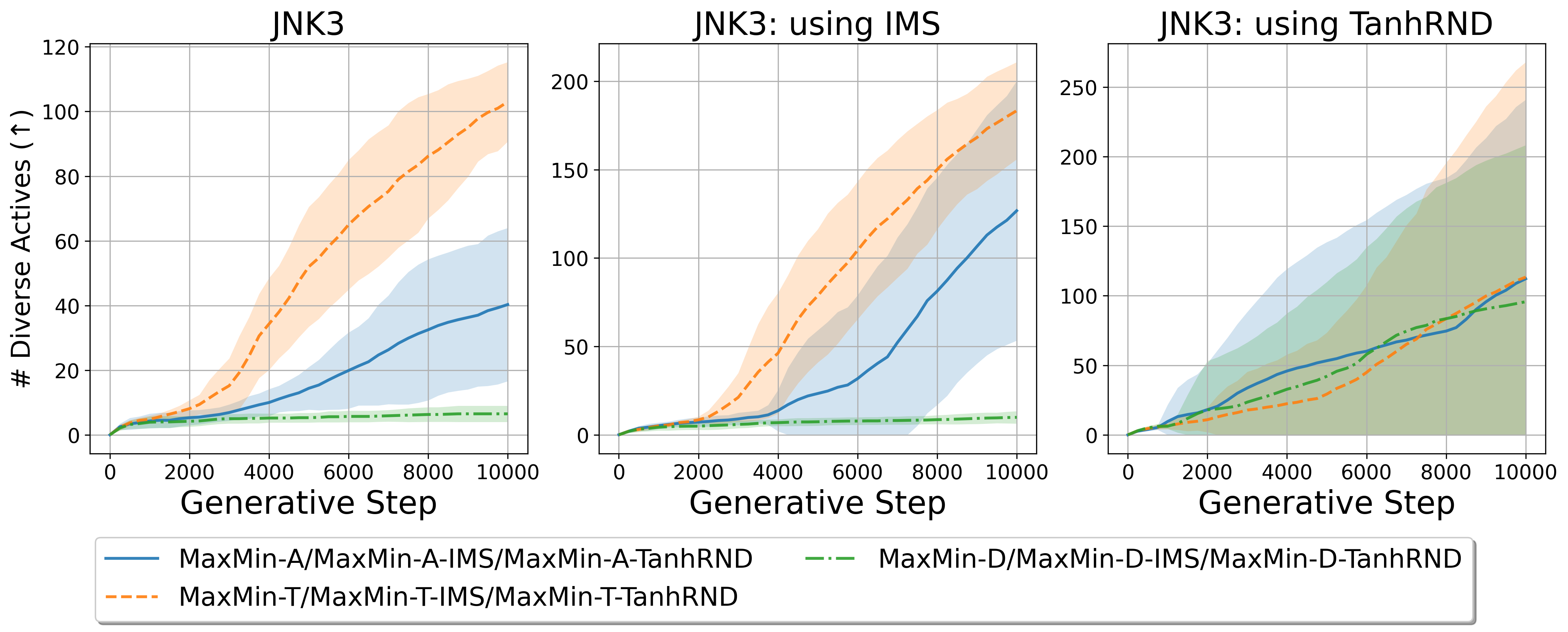}
         \caption{JNK3}
         \label{fig:jnk3_circles_maxmin}
     \end{subfigure}
        \caption{Total number of diverse activities after $g$ generative steps evaluated on reward functions based on the DRD2, GSK3$\beta$, or JNK3 predictive model. The total number of diverse actives is plotted for every 250th generative step.}
        \label{fig:circles_maxmin}
\end{figure}

\begin{figure}[t]
     \centering
     \begin{subfigure}[b]{0.47\textwidth}
         \centering
         \includegraphics[width=\textwidth]{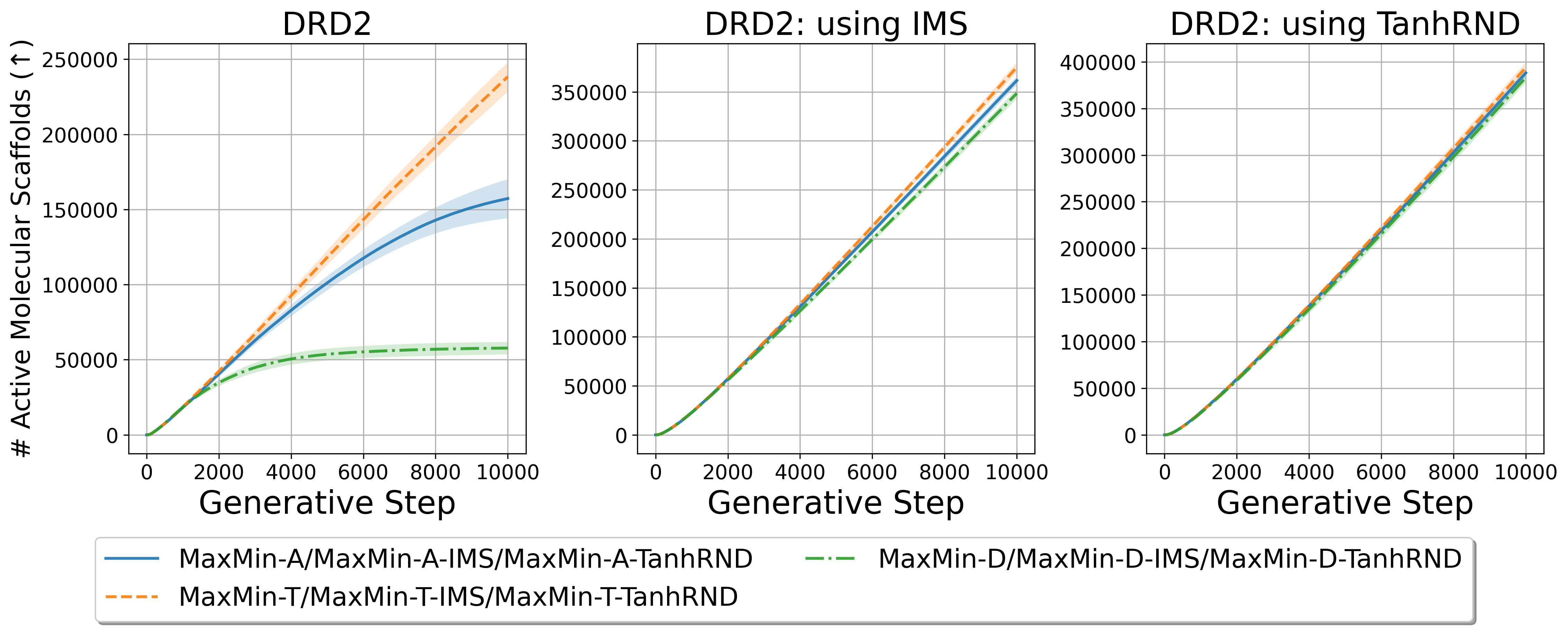}
         \caption{DRD2}
         \label{fig:drd2_scaffolds_maxmin}
     \end{subfigure}
     \hfill
     \begin{subfigure}[b]{0.47\textwidth}
         \centering
         \includegraphics[width=\textwidth]{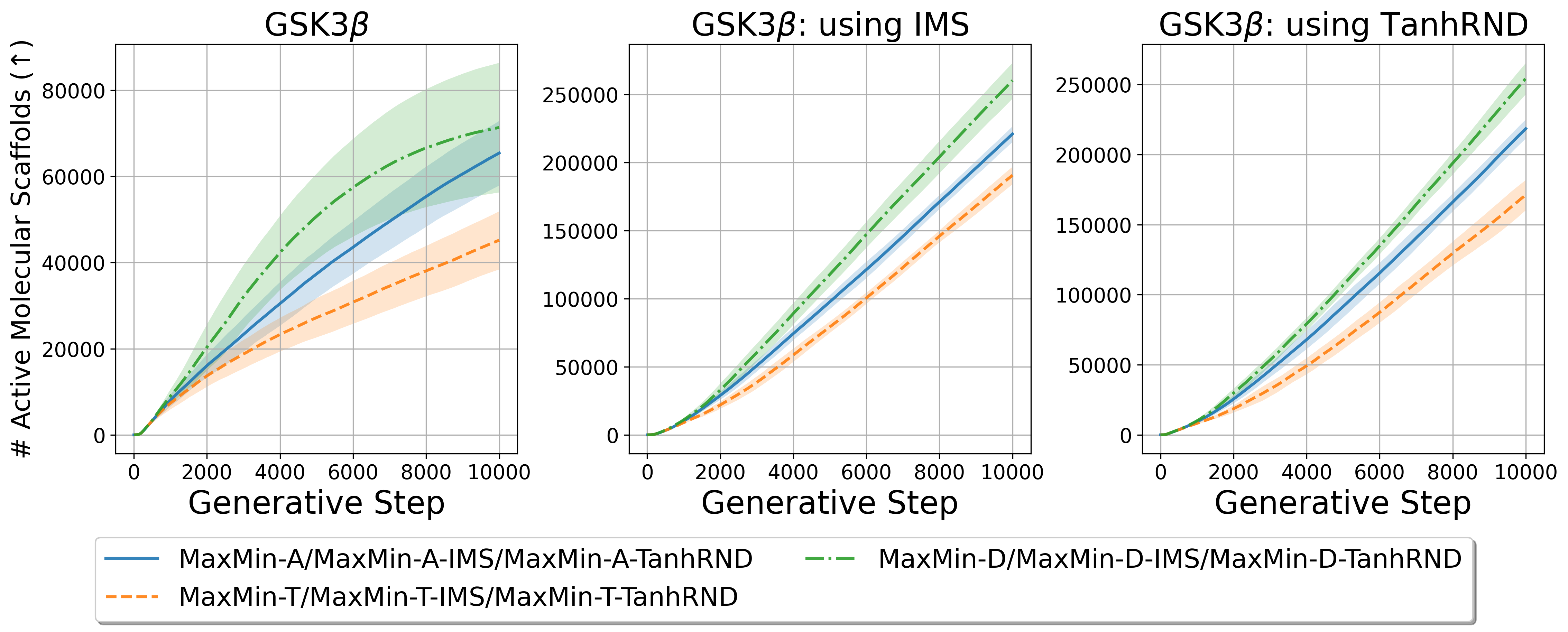}
         \caption{GSK3$\beta$}
         \label{fig:gsk3b_scaffolds_maxmin}
     \end{subfigure}
     \hfill
     \begin{subfigure}[b]{0.47\textwidth}
         \centering
         \includegraphics[width=\textwidth]{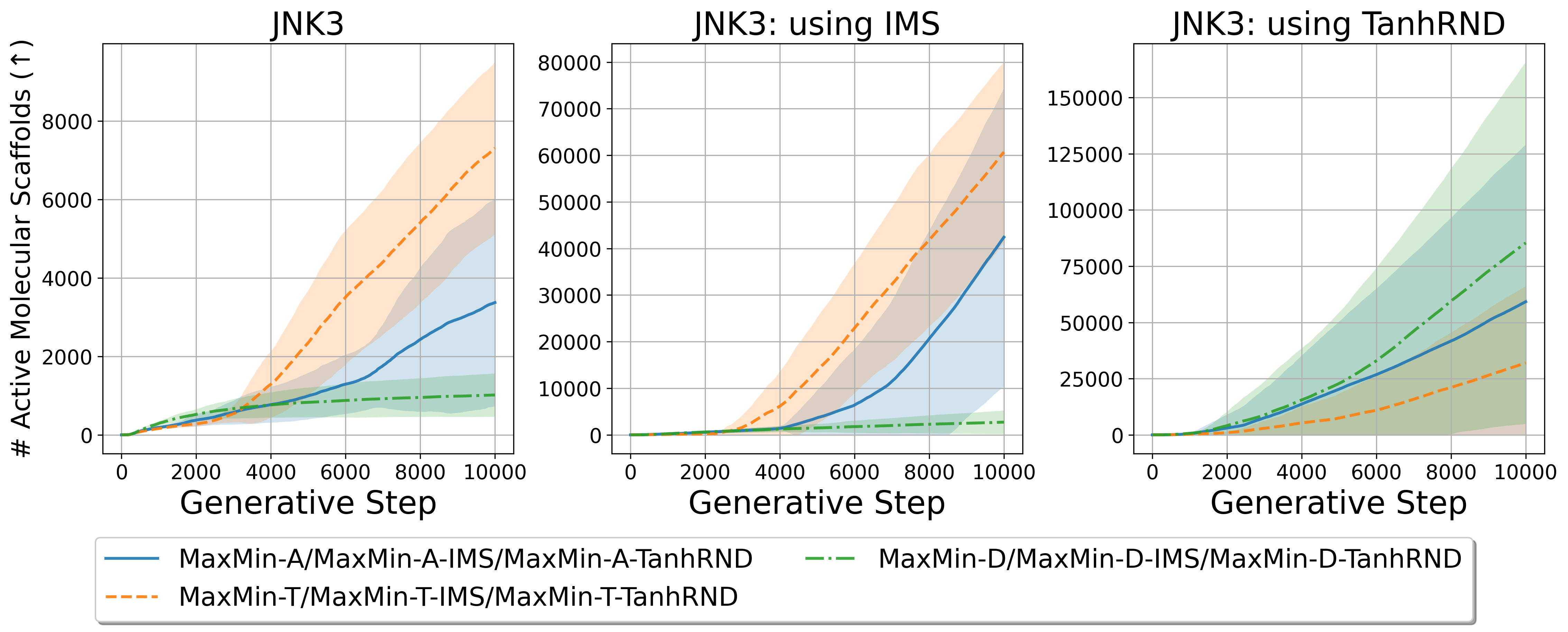}
         \caption{JNK3}
         \label{fig:jnk3_scaffolds_maxmin}
     \end{subfigure}
        \caption{Total number of molecular scaffolds after $g$ generative steps evaluated on reward functions based on the DRD2, GSK3$\beta$, or JNK3 predictive model.}
        \label{fig:scaffolds_maxmin}
\end{figure}

\Cref{fig:reward_maxmin} displays the extrinsic reward per generative step on the DRD2-, GSK3$\beta$-, and JNK3-based reward functions. For clarity of presentation, we display the moving averages with a window size of 101. Each line shows the average, while the shaded area shows the standard deviation.  For all configurations of the MaxMin algorithm, we mostly observe similar extrinsic rewards, but MaxMin-D sometimes displays lower and sometimes higher rewards. \Cref{fig:circles_maxmin} shows the total number of diverse actives up to the current generative step. For all experiments, except MaxMin-D-TanhRND on JNK3, MaxMin-T generated the largest number of diverse actives, while MaxMin-A is second-best. When using TanhRND on the JNK3-based reward function, all configurations display similar results, with high variability. \Cref{fig:scaffolds_maxmin} shows the total number of molecular scaffolds up to the current generative step. For the DRD2-based reward function, all configurations show similar trends when using IMS or TanhRND to enhance exploration, where MaxMin-T generates the largest number of scaffolds across all experiments. For the GSK3$\beta$-based reward function, MaxMin-T generates the smallest number of scaffolds, while MaxMin-D and MaxMin-A yield the largest and second largest number of scaffolds, respectively. On the JNK3 problem, MaxMin-A and MaxMin-T display similar trends when using IMS or no reward modification, yielding a larger number of scaffolds compared to MaxMin-D, which stagnates after a few thousand steps. When using TanhRND, all configurations display a large variability, where MaxMin-D yields the largest average and MaxMin-T the smallest average. 

Overall, MaxMin-T generates the largest number of diverse actives, while MaxMin-A illustrates comparable diversity and better diversity in terms of scaffolds. MaxMin-A better balances the two different diversity metrics and, therefore, we use this configuration in the main paper. 

\subsection{$k$-Medoids Clustering}
\begin{figure}[ht]
     \centering
     \begin{subfigure}[b]{0.47\textwidth}
         \centering
         \includegraphics[width=\textwidth]{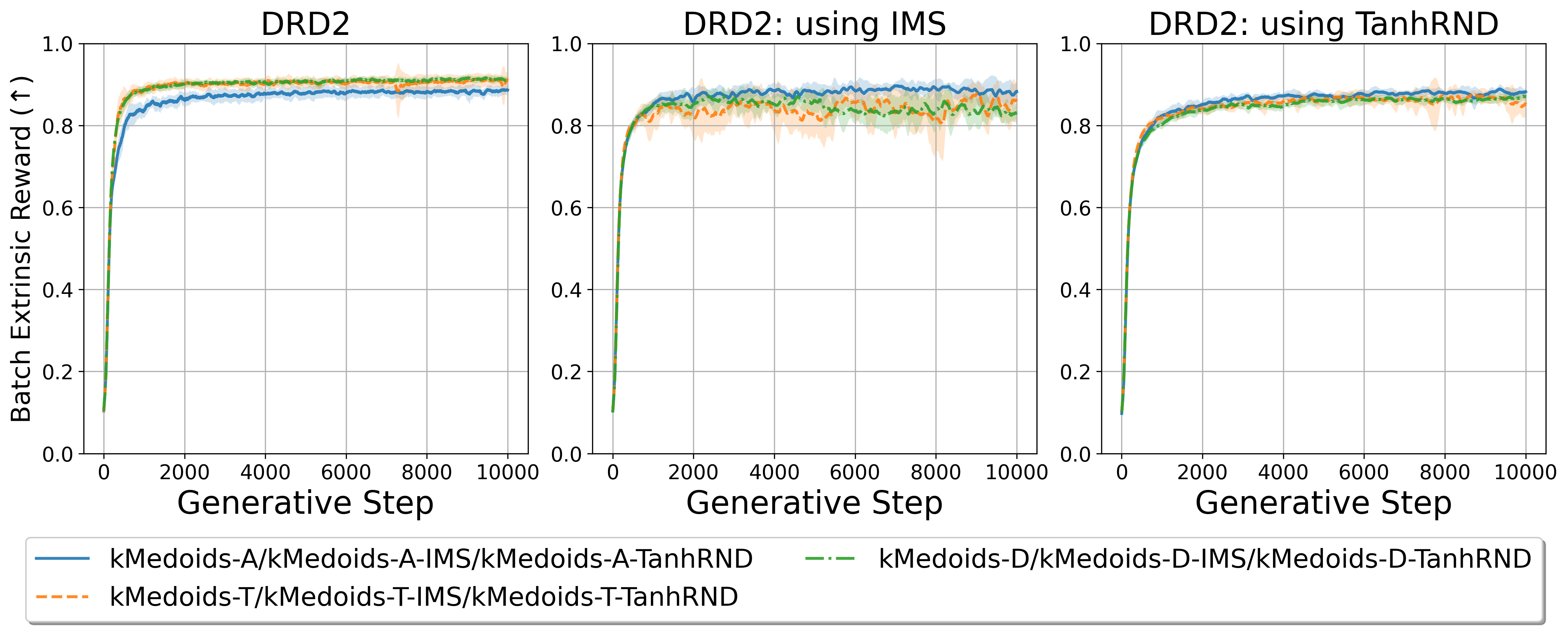}
         \caption{DRD2}
         \label{fig:drd2_reward_kmedoids}
     \end{subfigure}
     \hfill
     \begin{subfigure}[b]{0.47\textwidth}
         \centering
         \includegraphics[width=\textwidth]{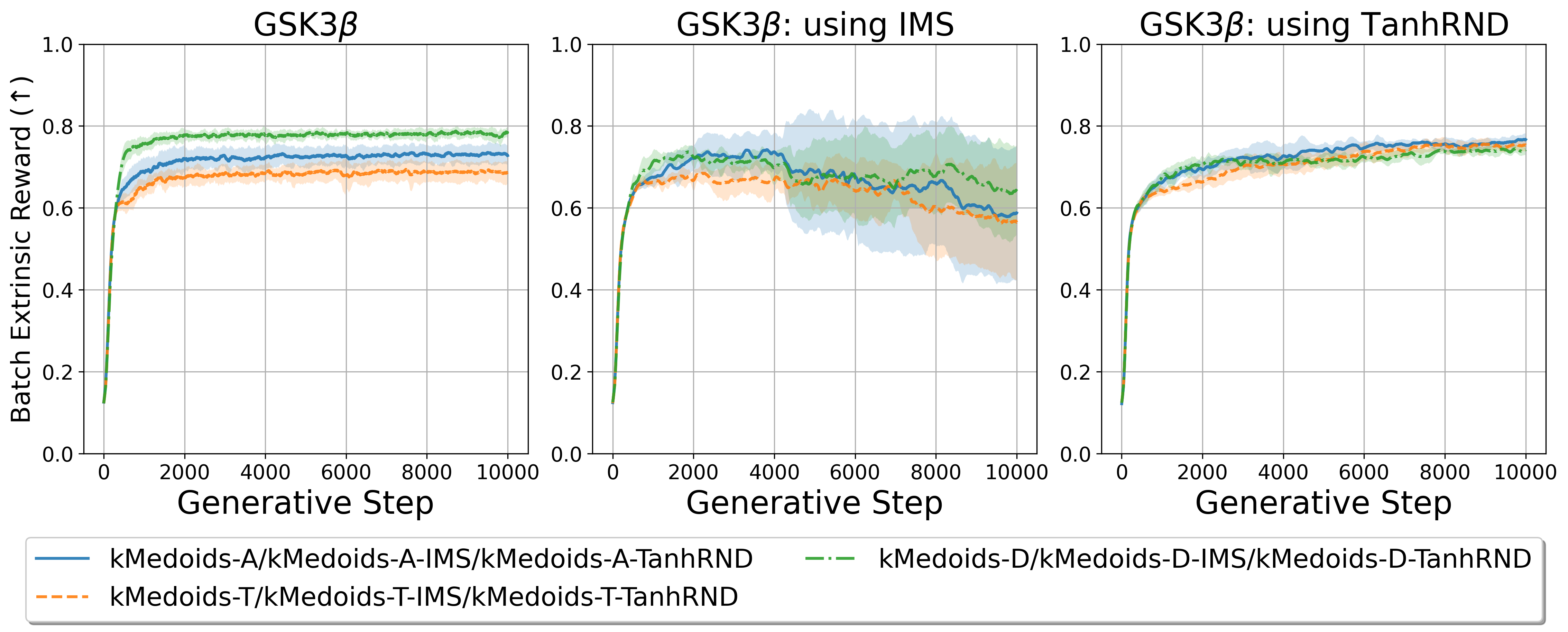}
         \caption{GSK3$\beta$}
         \label{fig:gsk3b_reward_kmedoids}
     \end{subfigure}
     \hfill
     \begin{subfigure}[b]{0.47\textwidth}
         \centering
         \includegraphics[width=\textwidth]{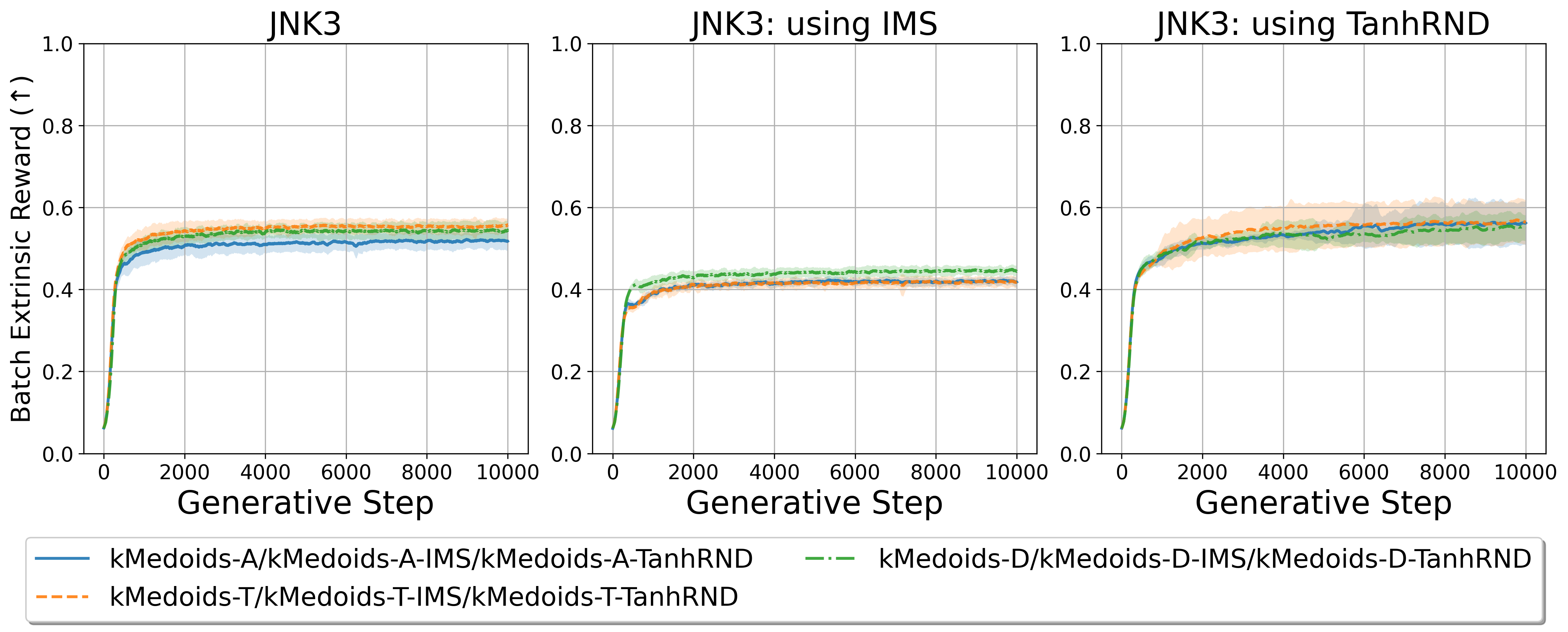}
         \caption{JNK3}
         \label{fig:jnk3_reward_kmedoids}
     \end{subfigure}
        \caption{Average extrinsic rewards per generative step across the mini-batch of SMILES evaluated on the DRD2-, GSK3$\beta$-, or JNK3-based reward functions. For clarity of presentation, we display the moving averages with a window size of 101. }
        \label{fig:reward_kmedoids}
\end{figure}

\begin{figure}[t]
     \centering
     \begin{subfigure}[b]{0.47\textwidth}
         \centering
         \includegraphics[width=\textwidth]{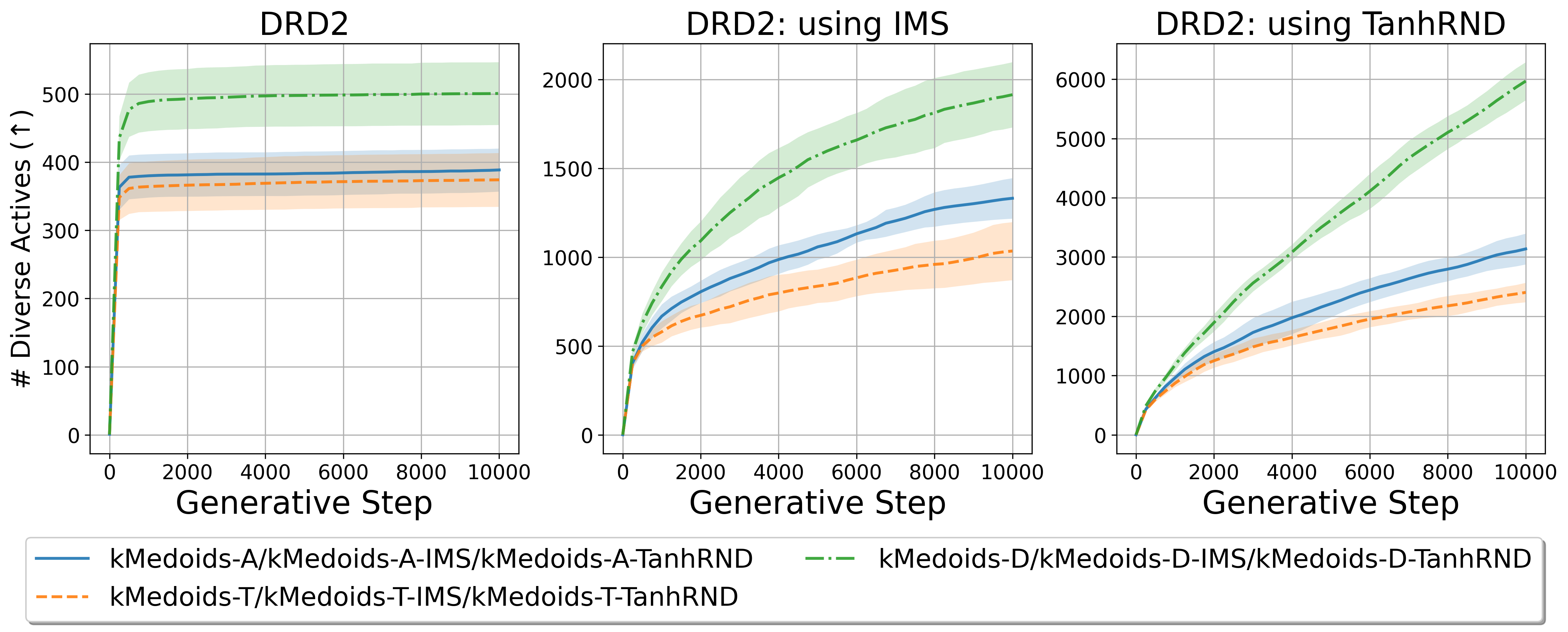}
         \caption{DRD2}
         \label{fig:drd2_circles_kmedoids}
     \end{subfigure}
     \hfill
     \begin{subfigure}[b]{0.47\textwidth}
         \centering
         \includegraphics[width=\textwidth]{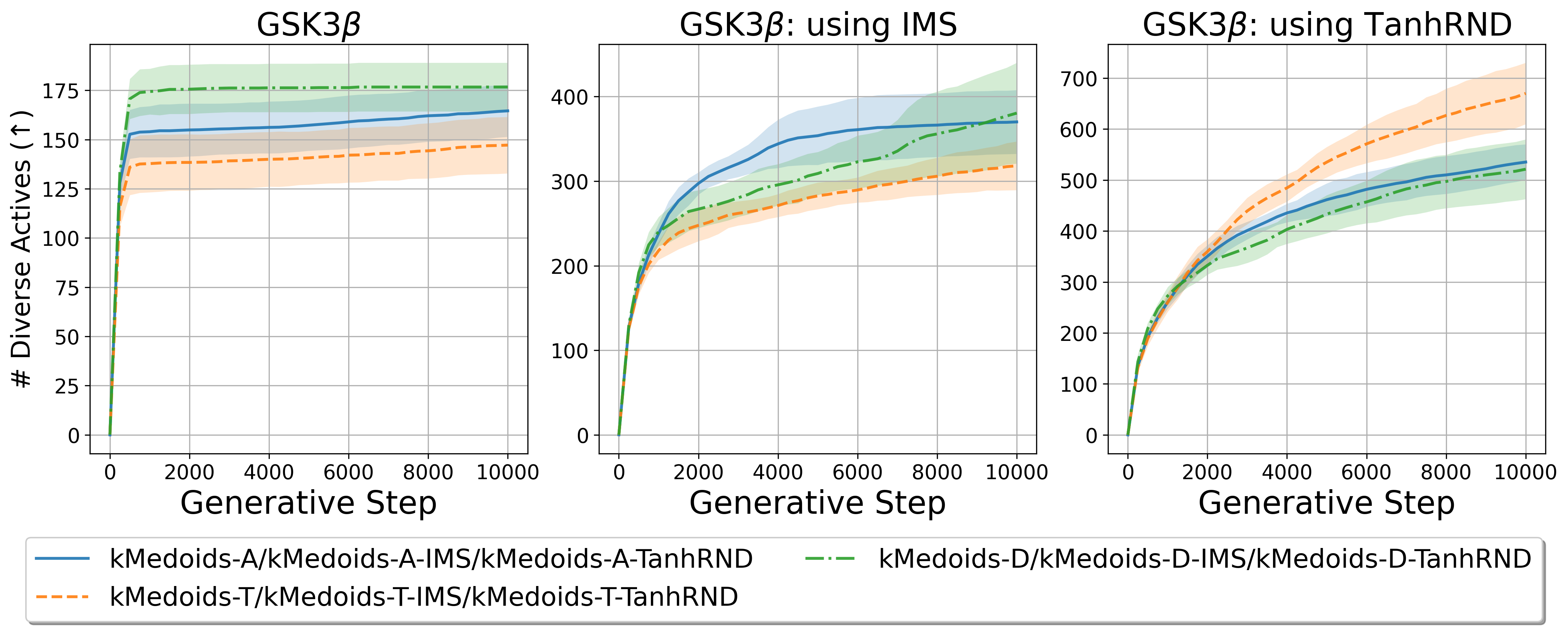}
         \caption{GSK3$\beta$}
         \label{fig:gsk3b_circles_kmedoids}
     \end{subfigure}
     \hfill
     \begin{subfigure}[b]{0.47\textwidth}
         \centering
         \includegraphics[width=\textwidth]{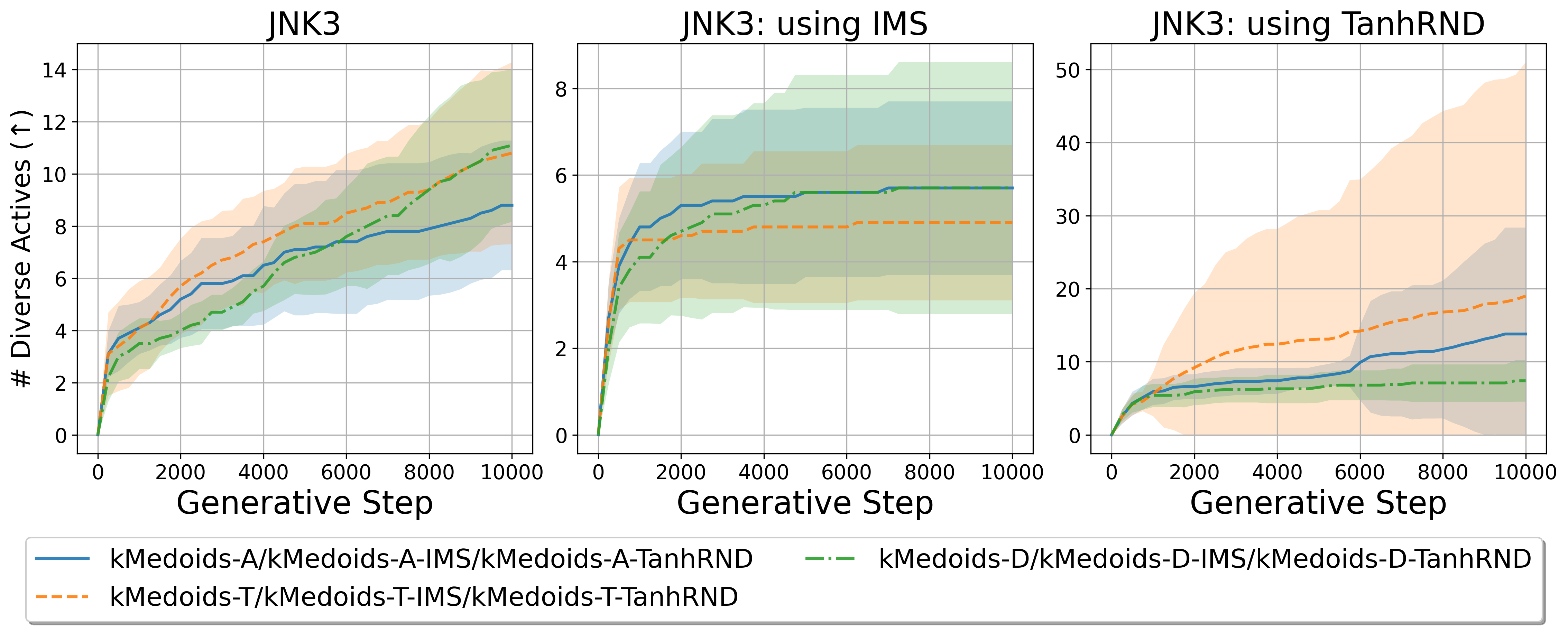}
         \caption{JNK3}
         \label{fig:jnk3_circles_kmedoids}
     \end{subfigure}
        \caption{Total number of diverse activities after $g$ generative steps evaluated on reward functions based on the DRD2, GSK3$\beta$, or JNK3 predictive model. The total number of diverse actives is plotted for every 250th generative step.} 
        \label{fig:circles_kmedoids}
\end{figure}

\begin{figure}[t]
     \centering
     \begin{subfigure}[b]{0.47\textwidth}
         \centering
         \includegraphics[width=\textwidth]{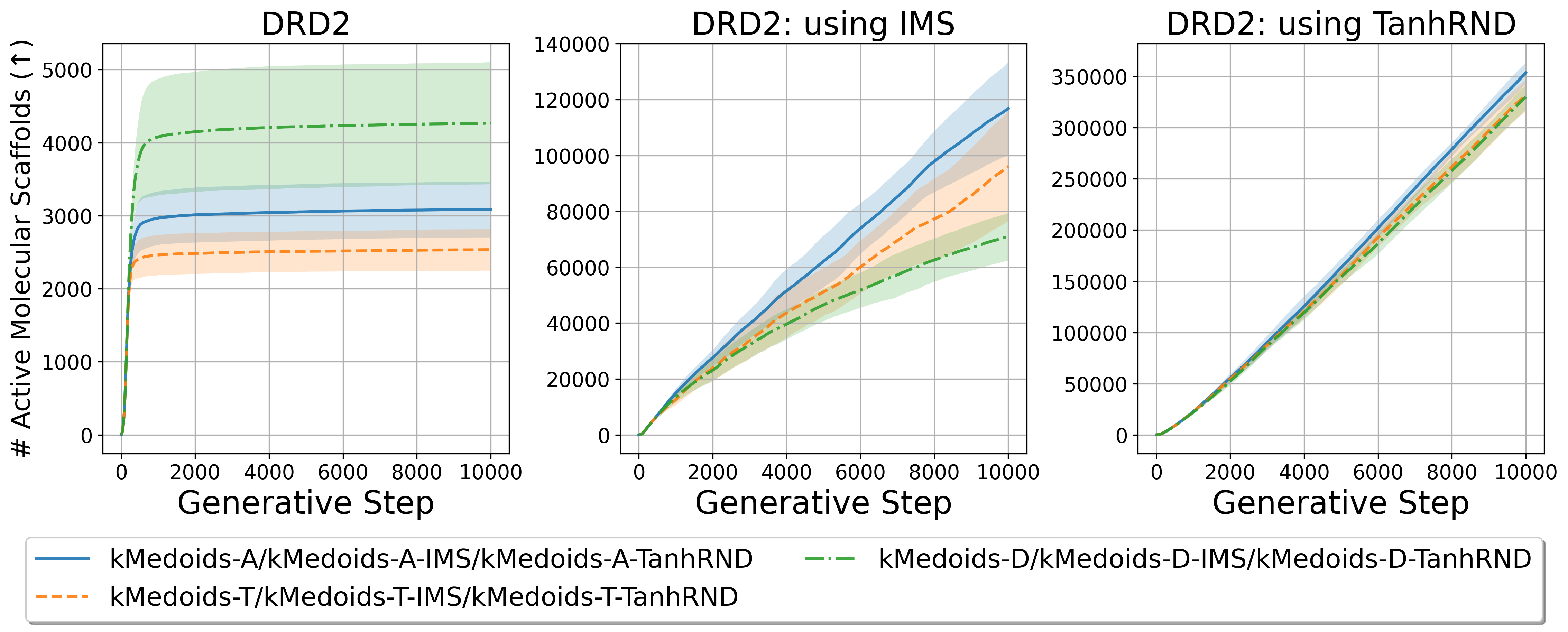}
         \caption{DRD2}
         \label{fig:drd2_scaffolds_kmedoids}
     \end{subfigure}
     \hfill
     \begin{subfigure}[b]{0.47\textwidth}
         \centering
         \includegraphics[width=\textwidth]{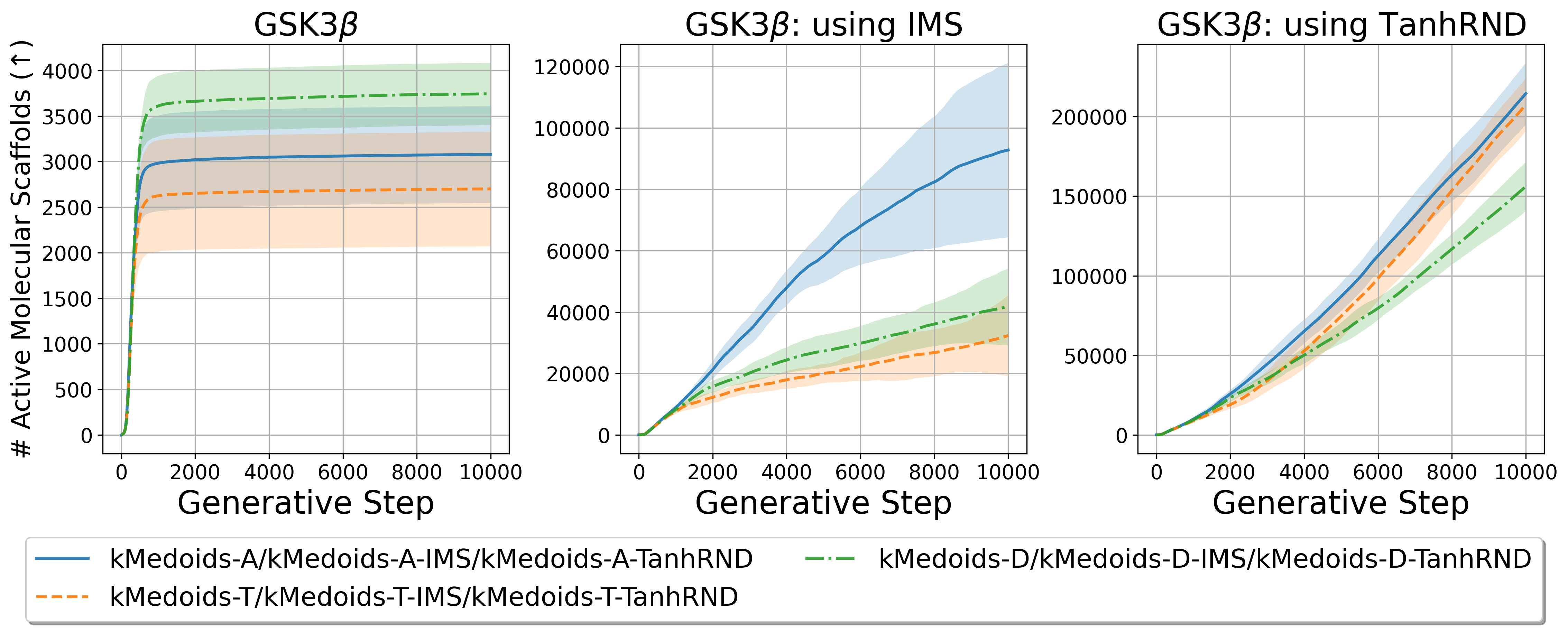}
         \caption{GSK3$\beta$}
         \label{fig:gsk3b_scaffolds_kmedoids}
     \end{subfigure}
     \hfill
     \begin{subfigure}[b]{0.47\textwidth}
         \centering
         \includegraphics[width=\textwidth]{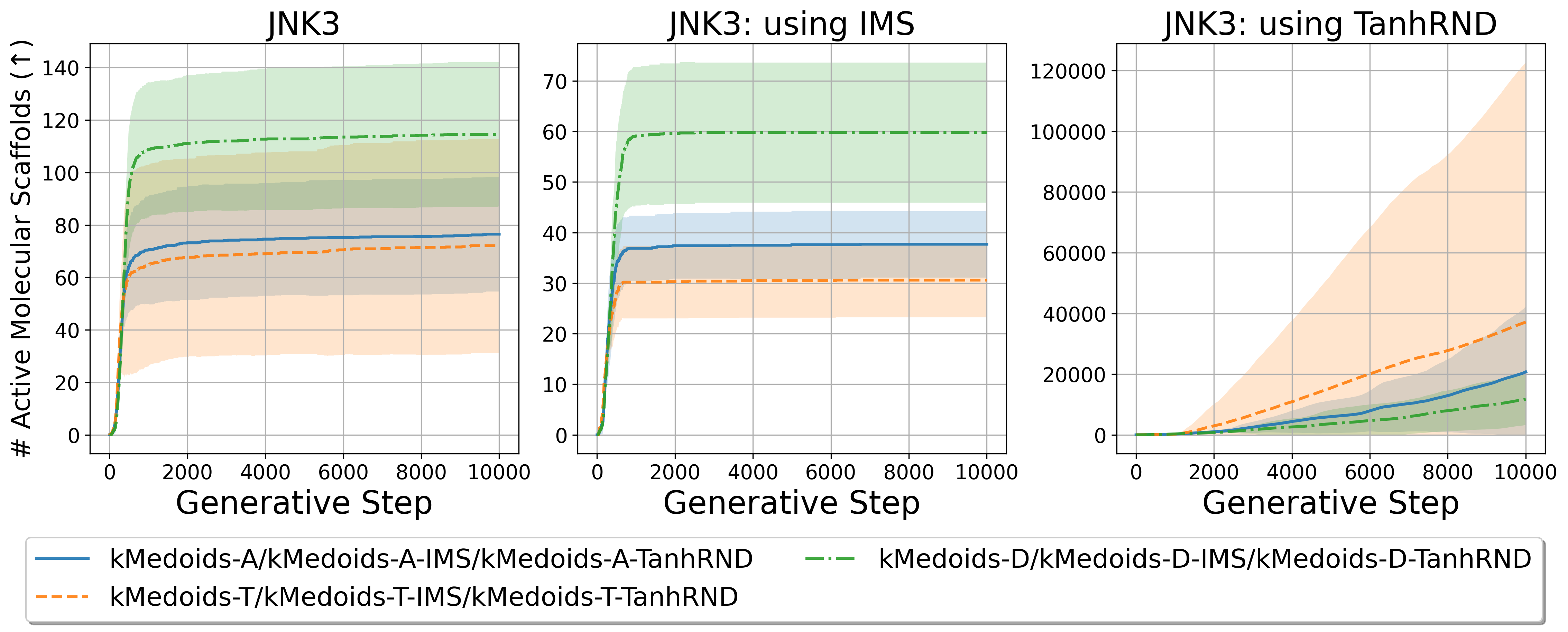}
         \caption{JNK3}
         \label{fig:jnk3_scaffolds_kmedoids}
     \end{subfigure}
        \caption{Total number of molecular scaffolds after $g$ generative steps evaluated on reward functions based on the DRD2, GSK3$\beta$, or JNK3 predictive model.}
        \label{fig:scaffolds_kmedoids}
\end{figure}

\Cref{fig:reward_kmedoids} displays the extrinsic reward per generative step on the DRD2-, GSK3$\beta$-, and JNK3-based reward functions when using $k$-medoids clustering for mini-batch diversification. For clarity of presentation, we display the moving averages with a window size of 101. Each line shows the average, while the shaded area shows the standard deviation.  For all different kernel matrices explored for $k$-medoids clustering, we observe similar trends. Rewards on the DRD2 problem are above 0.8,  rewards on GSK3$\beta$ are mostly between 0.8 and 0.6, and rewards on JNK3 are primarily below 0.6. \Cref{fig:circles_kmedoids} shows the total number of diverse actives up to the current generative step. On the DRD2-based reward function (see \cref{fig:drd2_circles_kmedoids}), kMedoids-D consistently yields the largest number of diverse actives, while kMedoids-A is slightly better than kMedoids-T. For the experiments on GSK3$\beta$ (see \cref{fig:gsk3b_circles_kmedoids}), kMedoids-T yields the largest number of diverse actives when using TanhRND, but otherwise generates a smaller number of diverse actives. kMedoids-A generates the second largest average number of diverse actives across all experiments, but its standard deviation overlaps with the other methods. For the JNK3-based reward function (see \cref{fig:jnk3_circles_kmedoids}), all methods generate a similar number of diverse actives. \Cref{fig:scaffolds_kmedoids} shows the total number of molecular scaffolds up to the current generative step. When modifying the reward (see middle and right plots in \Cref{fig:scaffolds_kmedoids}), the experiments of kMedoids-A generate the largest number of scaffolds, but their standard deviations overlap with the ones of kMedoids-T. When not modifying the extrinsic reward (see left plots in \Cref{fig:scaffolds_kmedoids}), fewer scaffolds are generated, where kMedoids-D performs the best. 

Overall, both kMedoids-D and kMedoids-T generate the largest number of diverse actives (for different reward functions), while kMedoids-A or kMedoids-T yield the largest number of scaffolds. We argue that kMedoids-A best balances the benefits of kMedoids-T and kMedoids-D, since it is always the second-best or best method. Therefore, we use this configuration in the main paper.

\section{Experimental Detatils}
\label{app:experimental_details}
\begin{algorithm}
\caption{Diverse Mini-Batch Selection for Drug Design}\label{alg:experiment} 
\begin{algorithmic}[1] 
\State \textbf{input:} $G,B,k,\theta_{prior},h$
\State $\mathcal{M} \gets \emptyset$ \Comment{Initialize memory}
\State $\theta \gets \theta_{\text{prior}}$ \Comment{The prior policy is fine-tuned}
 \For{$g=1,\dots,G$} \Comment{Generative steps}
 \State $\mathcal{L}(\theta) \gets 0$ 
 \State $\mathcal{K} \gets \emptyset$
    \For{$b=1,\dots,B$} \Comment{Large batch of SMILES}
        \State $t\gets 0$
        \State $a_t \gets a^{(\text{start})}$ \Comment{Start token is initial action}
        \State $s_{t+1} \gets a_{t}$
        \While{$s_{t+1}$ is not terminal}
            \State $t\gets t+1$
            \State $a_t \sim \pi_\theta(s_t)$
            \State $s_{t+1} \gets a_{0:t}$
        \EndWhile
        \State $\mathcal{B} \gets \mathcal{B} \cup s_{t+1}$
        \State Observe property score $r(s_{t+1})$
        \If{$r(s_{t+1}) \geq h$}
            \State $\mathcal{M} \gets \mathcal{M} \cup \{s_{t+1}\}$
        \EndIf  
        \State Compute and store penalty $f(s_{t+1}; \mathcal{M})$ 
    \EndFor
    \State Compute matrix kernel $L$ over $\mathcal{B}$
    \State Select $k$ SMILES from $\mathcal{B}$
    \For{$A\in Y$}
        \State Compute intrinsic reward $R_I(A; \mathcal{M})$ 
        \State Computed modified reward $\hat{R}(A)$
        \State Compute loss $\mathcal{L}_A (\theta)$ wrt $\hat{R}(A)$ 
        \State $\mathcal{L}(\theta) \gets \mathcal{L}(\theta) + \mathcal{L}_A (\theta)$
    \EndFor
    \State Update $\theta$ by minimizing $\mathcal{L}(\theta)$ in \cref{eq:loss}
\EndFor

 \State \textbf{output:} $\theta,\mathcal{M}$
\end{algorithmic} 
\end{algorithm}

The \textit{de novo} drug design problem can be modeled as a Markov decision process (MDP). Then, $a_t\in\mathcal{A}$ is the action taken at state $s_t$. We can define the current state as the sequence of performed actions up to round $t$
\begin{equation}
    s_t \coloneq a_{0:t-1} = a_0,a_1,\dots,a_{t-1},
\end{equation}
where the initial action is always the start token $a_0 = a^{\text{start}}$. This means that the distribution of the initial state $s_0$ is deterministic $p_0(s_0 = a^{\text{start}}) = 0$. The transition probabilities  are deterministic 
\begin{equation}
    P(s_{t+1}|s_t,a_t) = \delta_{ s_t  \mdoubleplus a_t},
\end{equation}
where $\mdoubleplus$ denotes the concatenation of two sequences. If action $a^{\text{stop}}$ is taken, the following state is terminal, stopping the current generation process and subsequently evaluating the generated molecule,
\begin{equation}
    P(\text{terminal}|s_t,a^{\text{stop}}) = 1,
\end{equation}
where $\delta_z$ denotes the Dirac distribution at $z$. The extrinsic reward epsidoci such that
\begin{equation}
    R(s_t,a_t) = R(a_{0:t}) =
    \begin{cases}
        r(s_{t+1}) & \text{if } a_t = a^{\text{stop}},\\
        0 & \text{otherwise,}
    \end{cases}
\end{equation} 
where reward $r(s_{T}) \in [0,1]$ (only observable at a terminal state) measures the desired property, which we want to optimize, of molecule $A = a_{1:T-2}$. We let $T$ denote the round that a terminal state is visited, i.e., $a_{T-1} = a^{\text{stop}}$. Note that in practice, the string between the start and stop tokens encodes a molecule such that $a_{1:T-2}$ is equivalent to $a_{0:T-1}$ during evaluation. The objective is to fine-tune a policy $\pi_\theta$, parameterized by $\theta$, to generate a structurally diverse set of molecules optimizing the property score $r(\cdot)$.

We use the REINVENT4 \cite{loeffler2024reinvent} framework to sequentially fine-tune the pre-trained (prior) policy. The algorithm is based on the \emph{augmented log-likelihood} defined by 
\begin{equation}
    \log \pi_{\theta_{\text{aug}}}(A) \coloneq \sum_{t=1}^{T-2}\log \pi_{\theta_{\text{prior}}}\left(a_t|s_{t}\right) + \sigma R(A),
\end{equation}
where $A=a_{1:T-2}$ is a generated molecule, $\sigma$ is a scalar value, $\pi_{\theta_{\text{prior}}}$ is the (fixed) pre-trained policy. The policy $\pi_\theta$ is optimized by minimizing the squared difference between the augmented log-likelihood and policy likelihood given a mini-batch $Y$ of $k$ SMILES
\begin{equation}
\label{eq:loss}
\begin{split}
    \mathcal{L}(\theta) = \frac{1}{k}  \sum_{a_{1:T-2}\in Y} \left(\log \pi_{\theta_{\text{aug}}}(a_{1:T-2}) \right. \\
    \left. - \sum_{t=1}^{T-2}\log \pi_{\theta}(a_t|s_t)\right)^2.
\end{split}%
\end{equation}
Previous work has shown that minimizing this loss function is equivalent to maximizing the expected return, as for policy gradient algorithms \cite{guo2024augmented}.

In practice, at each step $g$ of the generative process, $B$ full trajectories/episodes (until reaching a terminal state) are rolled out, to obtain a batch $\mathcal{B}$ of generated SMILES. Each token in the SMILES is sampled from the multinomial distribution induced by the policy's action probabilities. Subsequently, $k$-DPP, the MaxMin algorithm or $k$-medoids clustering is used to select a mini-batch of $k$ trajectories (SMILES) from $\mathcal{B}$. The modified reward $\hat{R}(A)$ for each molecule $A\in Y$ is observed by the agent and subsequently used for fine-tuning. The modified reward $\hat{R}(A)$ is computed using the penalty function $f(A)$ and/or intrinsic reward $R_I$ (depending on which reward function is used). The penalty functions and intrinsic rewards use a bucket size of $M$ to determine the desired number of generated molecules with the same scaffold (we refer to \cite{blaschke2020memory,svensson2024diversityawarereinforcementlearningnovo} for more details). The modified reward is used to compute the loss $\mathcal{L}(\theta)$ in \cref{eq:loss}, i.e., we let $R(A) = \hat{R}(A)$ if the extrinsic reward is modified, e.g., via intrinsic reward or reward penalty. The policy parameters $\theta$ are updated with respect to the $\mathcal{L}(\theta)$ using a learning rate $\alpha$. \Cref{alg:experiment} illustrates the specific procedure utilized for \textit{de novo} drug design. The source code is available in a GitHub repository.\footnote{\url{https://github.com/hampusgs/diverse-mini-batch-selection-rl}}
Fine-tuning of the policy network is done on a single NVIDIA A40 GPU with 48GB RAM or NVIDIA T4 GPU with 16GB RAM using PyTorch 1.12.1 and CUDA 11.3 on a Linux-based system. We use the DPPy package \cite{GPBV19} with version 0.3.3 to perform exact sampling from $k$-DPP, using the default random seed. For $k$-medoids clustering, we use the FasterPAM algorithm \cite{schubert2021fast} from the kmedoids package \cite{schubert2022fast} with version 0.5.3.1. We use random initialization of medoids and at most 100 iterations. We use the MaxMin algorithm implemented by RDKit \cite{landrum2006rdkit} with version 2023.9.6.

\subsection{Reward Function}
\label{app:reward_function}
\begin{table*}[h]
    \setlength{\tabcolsep}{1mm}
    \centering
    \begin{tabular}{ccccccccc} \toprule
       Component & Weight & Transform type & high & low & $c_{\text{div}}$ & $c_{\text{si}}$ & $c_{\text{se}}$ & $k$ \\ \midrule
        Molecular weight & 1 & Double sigmoid & 550 & 200 & 500 & 20 & 20 & \textendash \\ 
        \# hydrogen bond doners & 1 & Reverse sigmoid & 6 & 2 & \textendash & \textendash & \textendash & 0.5\\
        QED & 1 & None & \textendash & \textendash & \textendash & \textendash & \textendash  & \textendash \\
        Custom Alerts & 1 & None & \textendash  & \textendash & \textendash & \textendash & \textendash& \textendash \\
        Predictive oracle & 5 & None & \textendash & \textendash & \textendash & \textendash & \textendash & \textendash \\ \bottomrule
    \end{tabular}
    \caption{Parameters for scoring components in the REINVENT4 \cite{loeffler2024reinvent} framework. A geometric mean is used to combine them into the extrinsic reward observed by the agent.}
    \label{tab:components}
\end{table*}
Our experiments utilize scoring components of REINVENT4 \cite{loeffler2024reinvent} to define the extrinsic reward using a geometric mean. In addition, we implement a scoring component using the predictive oracles of the Dopamine Receptor D2 (DRD2), Glycogen Synthase Kinase 3 Beta (GSK3$\beta$) and c-Jun N-terminal Kinases-3 (JNK3) oracles from TD Commons \cite{Huang2021tdc,Velez-Arce2024tdc}. The weight and parameters for each scoring component are displayed in \cref{tab:components}. Predictive oracle functions, providing the activity values, are provided by PyTDC 1.1.4. Fingerprints and QED are computed using RDKit 2023.9.6. For the custom alerts, we use the following default chemical patterns in the SMARTS language:
\begin{itemize}
    \item \phantom{w }[*;r8]
    \item \phantom{w }[*;r9]
    \item \phantom{w }[*;r10]
    \item \phantom{w }[*;r11]
    \item \phantom{w }[*;r12]
    \item \phantom{w }[*;r13]
    \item \phantom{w }[*;r14]
    \item \phantom{w }[*;r15]
    \item \phantom{w }[*;r16]
    \item \phantom{w }[*;r17]
    \item \phantom{w }[\#8][\#8] 
    \item \phantom{w }[\#6;+]
    \item \phantom{w }[\#16][\#16]
    \item \phantom{w }[\#7;!n][S;!\$(S(=O)=O)]
    \item \phantom{w }[\#7;!n][\#7;!n]
    \item \phantom{w }C\#C
    \item \phantom{w }C(=[O,S])[O,S] 
    \item \phantom{w }[\#7;!n][C;!\$(C(=[O,N])[N,O])][\#16;!s] 
    \item \phantom{w }[\#7;!n][C;!\$(C(=[O,N])[N,O])][\#7;!n]
    \item \phantom{w }[\#7;!n][C;!\$(C(=[O,N])[N,O])][\#8;!o]
    \item \phantom{w }[\#8;!o][C;!\$(C(=[O,N])[N,O])][\#16;!s]
    \item \phantom{w }[\#8;!o][C;!\$(C(=[O,N])[N,O])][\#8;!o] 
    \item \phantom{w }[\#16;!s][C;!\$(C(=[O,N])[N,O])][\#16;!s]
\end{itemize}

\subsection{Hyperparameters}
\Cref{tab:hyperparameters} displays the hyperparameters used in the experimental evaluation. We run for $G=10000$ generative/training steps to investigate the chemical exploration over a large number of steps. We generate a large set $\mathcal{B}$ of $|\mathcal{B}| = B = 640 $ instances/items since we argue that ten times the number of items we want to choose (i.e., $k$) is sufficient to generate diverse solutions. This is supported by our experiments. We use a distance threshold $D=0.7$ as suggested by \cite{renz2024diverse} since there is a significant decrease in the probability of similar bioactives beyond this threshold~\cite{10.12688/f1000research.8357.2}. When computing the diversity in terms of both scaffolds and diverse actives, we only regard active molecules, defined as molecules with both QED and predicted activity larger than $h = 0.5$. The activity models are trained on binary classification tasks, such that a value larger than $h = 0.5$ means that the molecule is most likely to be active. A QED of $0.5$ is close to the mean QED of approved drugs \cite{bickerton2012quantifying}. Otherwise, we use the default hyperparameters of REINVENT4 \cite{loeffler2024reinvent}.

\begin{table*}[ht]
    \centering
    \begin{tabular}{c|c} \toprule
       Parameter & Value  \\ \midrule
        $B$ & 640 \\ 
        $G$ & \num{10000}\\
        $h$ & 0.5 \\
        $k$ & 64 \\
        $D$ & 0.7\\
        $\sigma$ & 128\\
        $\alpha$ & \num{0.0001}\\
        Optimzer & Adam \cite{kingma2017adammethodstochasticoptimization}\\
        $M$ & 25\\
        $|\mathcal{A}|$ & 34 \\ \bottomrule
    \end{tabular}
    \caption{Hyperparameters for the experimental evaluation using the REINVENT4 \cite{loeffler2024reinvent} framework.}
    \label{tab:hyperparameters}
\end{table*}

\section{Analysis of predictive activity models}
\label{app:analysis_activity_models}
\begin{figure*}[ht]
\centering
\begin{subfigure}{0.47\textwidth}
\centering
    \includegraphics[width=\textwidth]{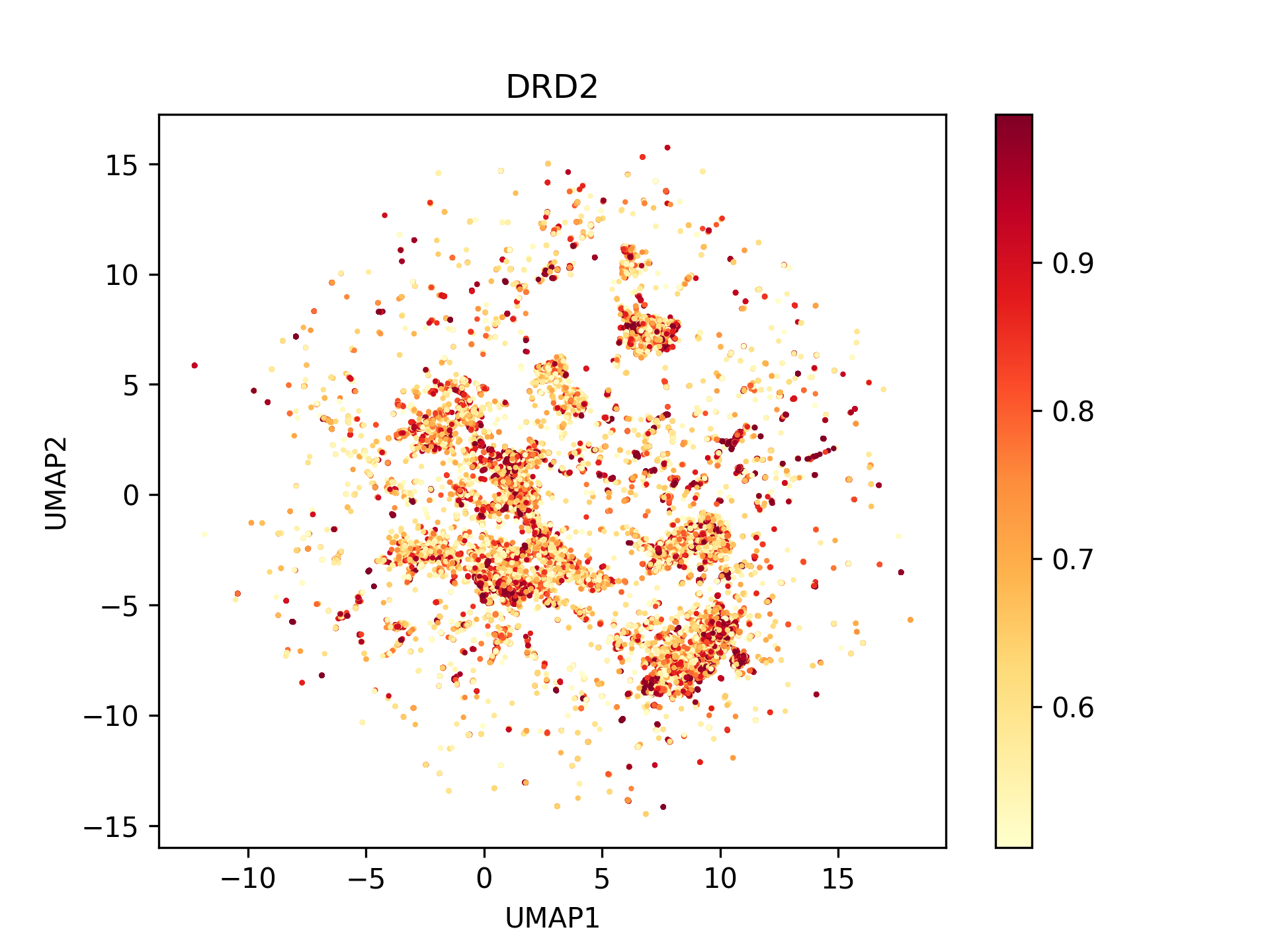}
    \caption{DRD2}
    \label{fig:umap_drd2}
\end{subfigure}
\hfill
\begin{subfigure}{0.47\textwidth}
\centering
    \includegraphics[width=\textwidth]{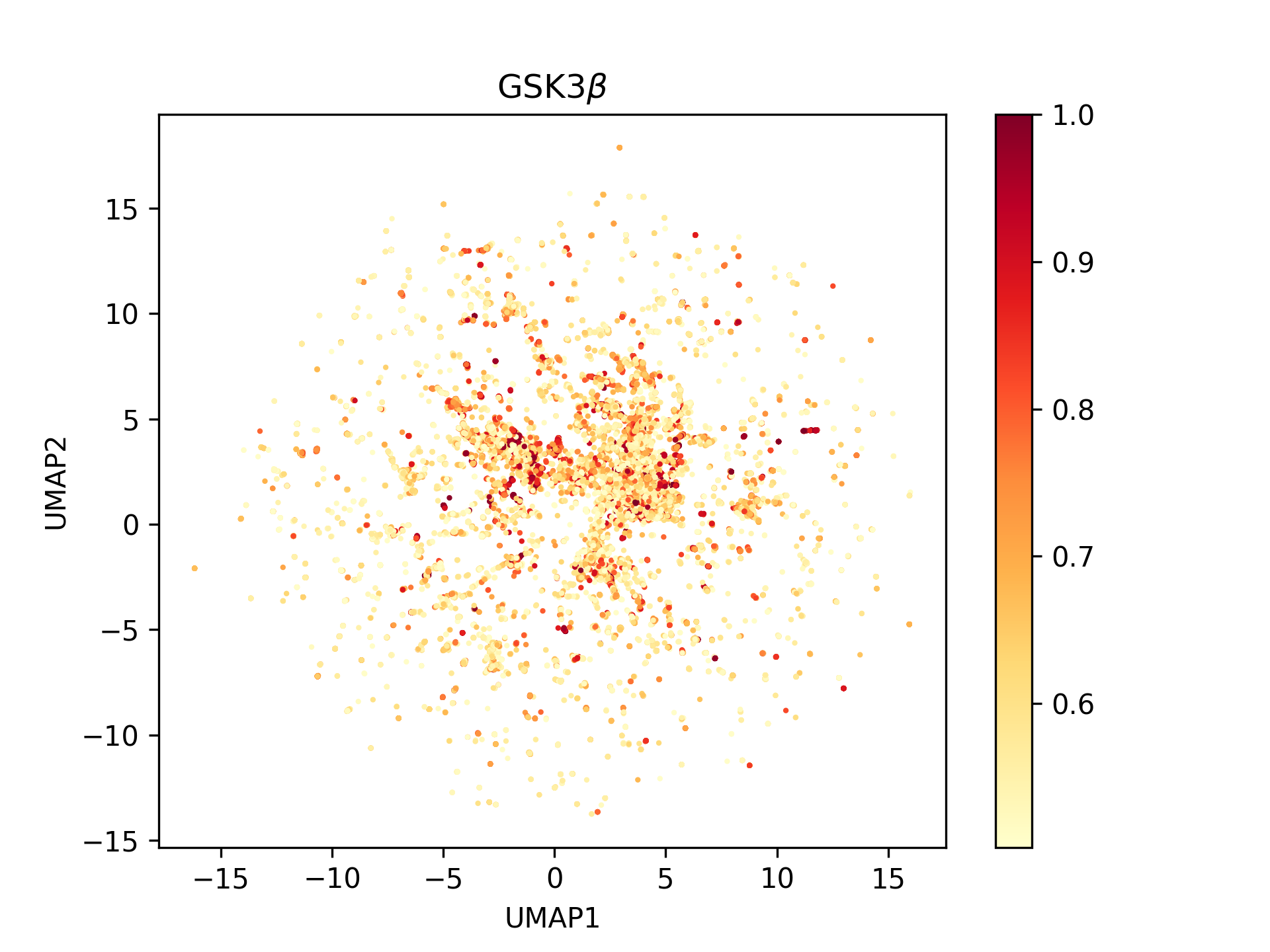}
    \caption{GSK3$\beta$}
    \label{fig:umap_gsk3b}
\end{subfigure}
\hfill
\begin{subfigure}{0.47\textwidth}
\centering
    \includegraphics[width=\textwidth]{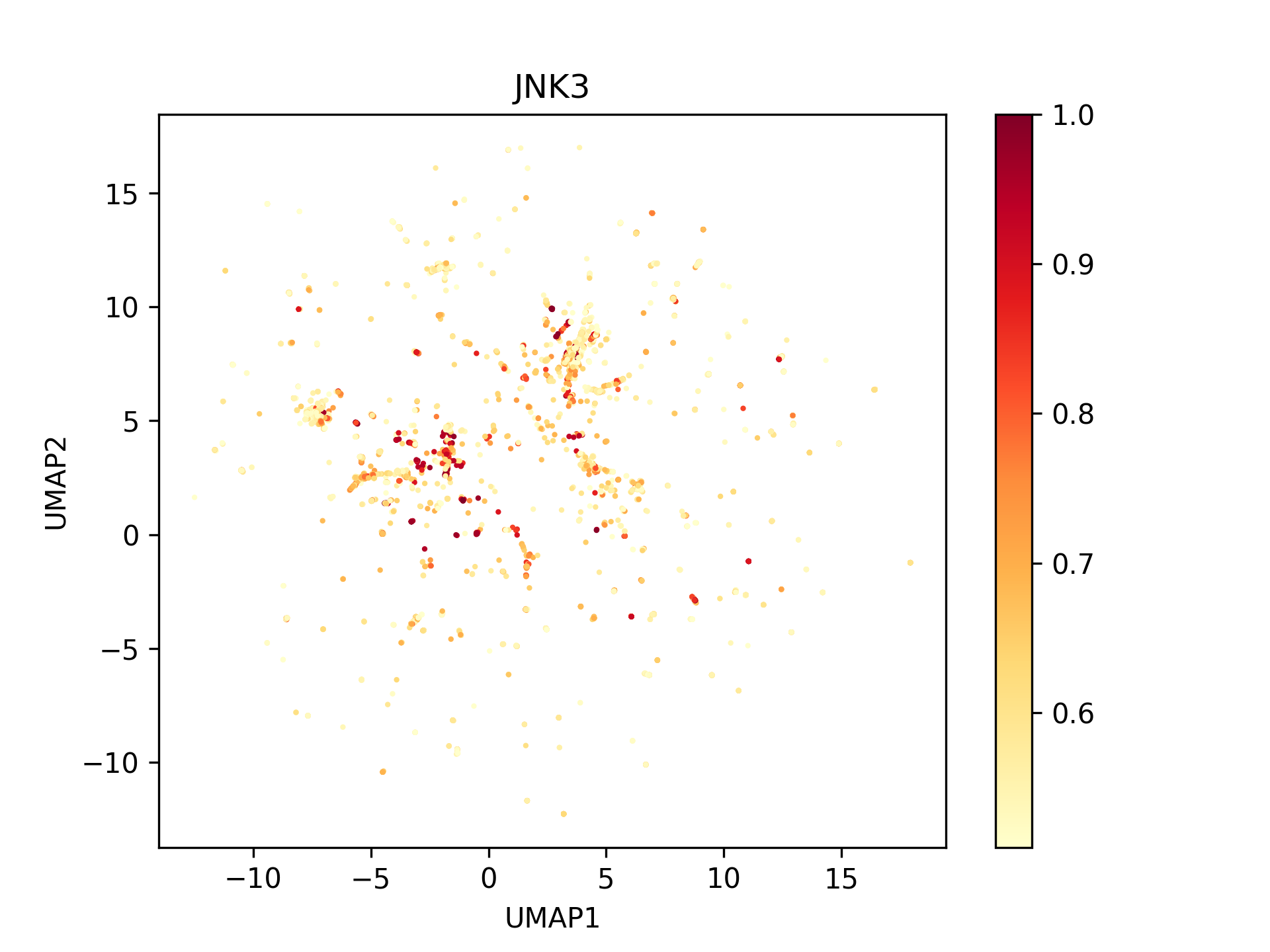}
    \caption{JNK3}
    \label{fig:umap_jnk3}
\end{subfigure}
        
\caption{2-dimensional UMAP projection of 200 PCA features. The PCA features are derived from 2048-bits ECFP (Morgan) fingerprints. We only display the active molecules with a predicted activity of more than 0.5.}
\label{fig:umap}
\end{figure*}

To better understand the underlying reward space, we visualize the molecules from ChEMBL25 \cite{gaulton2017chembl}, in total 2474589 molecules, on the three predictive activity models (oracles) investigated in this work. ECFP (Morgan) fingerprints with 2048 bits are reduced to 200 features using principal components analysis (PCA) using scikit-learn \cite{sklearn_api}. These 200 features are subsequently reduced to 2 dimensions using UMAP \cite{mcinnes2018umap}. For clarity, we only display the active molecules with a predicted activity of more than 0.5. Only fingerprints are used for the different predictive models to match the fingerprints used for the corresponding predictive models in Therapeutics Data Commons \cite{Huang2021tdc,Velez-Arce2024tdc}.

For DRD2 oracle, ECFPC3 fingerprints (using counts and features) are calculated using RDKit \cite{landrum2006rdkit}  and visualized in \cref{fig:umap_drd2}. There are 58843 active molecules in total for the DRD2 oracle. For the GSK3$\beta$ oracle, ECFP2 fingerprints are calculated using RDKit \cite{landrum2006rdkit} and visualized in \cref{fig:umap_gsk3b}. There are 44066 active molecules in total for the GSK3$\beta$ oracle. For the JNK3 oracle, ECFP2 fingerprints are calculated using RDKit \cite{landrum2006rdkit} and visualized in \cref{fig:umap_jnk3}. There are 7249 active molecules in total for the JNK3 oracle.